\documentclass[letterpaper, 10 pt, conference]{ieeeconf} 

\usepackage{times}

\usepackage{multicol}
\usepackage[bookmarks=true]{hyperref}

\usepackage{graphicx}
\usepackage{bm}
\usepackage{url}
\usepackage{xspace}
\usepackage{amsfonts}
\usepackage{algpseudocode}
\usepackage{subcaption}
\usepackage{amsmath}
\usepackage{amssymb}
\usepackage{mathtools}

\usepackage[ruled, linesnumbered]{algorithm2e}
\newcommand{\eqdef}{\overset{\mathrm{def}}{=\joinrel=}}

\usepackage{comment}
\newcount\Comments
\usepackage{color}
\definecolor{darkgreen}{rgb}{0,0.7,0}
\definecolor{gray}{rgb}{0.95,0.95,0.95}
\definecolor{lilach}{rgb}{0.95,0.5,0.95}
\definecolor{purple}{rgb}{1,0,1}
\definecolor{teal}{rgb}{0.2,0.95,0.95}
\definecolor{orange}{rgb}{1,0.5,0}
\newcommand{\kibitz}[2]{\ifnum\Comments=1{\color{#1}{#2}}\fi}
\newcommand{\sk}[1]{\kibitz{purple}{[sk: #1]}}

\newcommand{\dd}[1]{\kibitz{darkgreen}{[david: #1]}}

\newtheorem{definition}{Definition}

\DeclareMathOperator*{\argmax}{arg\,max}

\newcommand{\VOA}{VOA\xspace}

\newcommand{\VOAFunc}{\Utility^{VOA}}

\newcommand{\Utility}{U}

\newcommand{\pose}{p}
\newcommand{\objPose}{\pose}
\newcommand{\poses}{\mathcal{P}}

\newcommand{\objStablePose}{\pose}
\newcommand{\allPoses}{\poses}
\newcommand{\allObjStablePoses}{\allPoses}
\newcommand{\objStablePoses}{\allPoses}
\newcommand{\objStablePosesSampled}{\tilde{\allPoses}}
\newcommand{\config}{x}
\newcommand{\configs}{\mathcal{X}}

\newcommand{\graspConfig}{g}
\newcommand{\graspConfigs}{\mathcal{G}}
\newcommand{\graspConfigSet}{\graspConfigs_{\actorSymbol}}
\newcommand{\graspConfigMax}{\graspConfig^{max}}

\newcommand{\helperConfig}{\config}
\newcommand{\helperConfigs}{\tilde{\configs}}
\newcommand{\allHelperConfigs}{\configs}
\newcommand{\observation}{o}
\newcommand{\observations}{O}
\newcommand{\allObservations}{\observations}
\newcommand{\estimatedObservation}{\hat{\observation}} 
 
\newcommand{\observationPerceivedProbability}{\hat{P}}
\newcommand{\sensorFunc}{\mathcal{O}}
\newcommand{\estimatedSensorFunc}{\hat{\sensorFunc}}

\newcommand{\belief}{\beta}
\newcommand{\beliefs}{\mathcal{B}}
\newcommand{\poseBelief}{\belief}

\newcommand{\poseBeliefs}{\beliefs}
\newcommand{\allPoseBeliefs}{\poseBeliefs}
\newcommand{\beliefUpdateFunc}{\tau}
\newcommand{\graspScoreFunc}{\gamma}
\newcommand{\updatedBelief}{\poseBelief^{\observation,\helperConfig}}

\newcommand{\predictedUpdatedBelief}{\poseBelief^{\predictedObs,\helperConfig}}
\newcommand{\predictedUpdatedBeliefDet}{\poseBelief^{\predictedObs}}

\newcommand{\graspScoreExpectation}{\bar{\graspScoreFunc}}

\newcommand{\actor}{actor}
\newcommand{\helper}{helper}

\newcommand{\expectedGraspScore}{expected grasp score}
\newcommand{\graspScore}{grasp score}
\newcommand{\observationprediction}{observation prediction}

\newcommand{\graspConfiguration}{grasp configuration}

\newcommand{\obsSimScore}{\omega}
\newcommand{\predictedObs}{\hat{\observation}}

\newcommand{\VDFunc}{\Utility^{VD}}

\newcommand{\predictedPoseBelief}{\hat{\poseBelief}}
\newcommand{\poseBeliefActor}{\poseBelief_{\actorSymbol}}
\newcommand{\poseBeliefHelper}{\poseBelief_{\helperSymbol}}

\newcommand{\actorBeliefUpdate}{\beliefUpdateFunc_{\actorSymbol}}

\newcommand{\assistanceAction}{\helperConfig}

\newcommand{\helperSymbol}{h}
\newcommand{\actorSymbol}{a}
\newcommand{\helperBelief}{\belief_{\helperSymbol}}
\newcommand{\actorBelief}{\belief_{\actorSymbol}}

\newcommand{\predictedActorBelief}{\hat{\belief}_{\actorSymbol}}

\newcommand{\SCOREDIFF}{\delta}
\newcommand{\SCOREDIFFEXP}{\bar{\delta}}
\newcommand{\BESTSCOREEXPRATIO}{\delta^*}
\newcommand{\ADVENTAGE}{\mathcal{A}}
\newcommand{\CONFIDENCEDIFF}{c}

\newcommand{\lidar}{lidar}

\usepackage{array}
\usepackage{booktabs}
\usepackage{multirow}
\usepackage{makecell}
\usepackage{caption}
\usepackage{adjustbox}

\title{\LARGE \bf
Value of Assistance for Grasping
}

\author{{Mohammad Masarwy, Yuval Goshen, David Dovrat and Sarah Keren} 
\thanks{*The authors are from the Taub Faculty of Computer Science, Technion - Israel Institute of Technology, Israel}
}

\begin{document}

\maketitle
\thispagestyle{empty}
\pagestyle{empty}

\begin{abstract}
    
In multiple realistic settings, a robot is tasked with grasping an object without knowing its exact pose and relies on a probabilistic estimation of the pose to decide how to attempt the grasp. We support settings in which it is possible to provide the robot with an observation of the object before a grasp is attempted but this possibility is limited and there is a need to 
decide which sensing action would be most beneficial. We support this decision by offering a novel {\em Value of Assistance} (\VOA) measure for assessing the expected effect a specific 
 observation will have on the robot's ability to complete it's task. 
We evaluate our suggested measure in simulated and real-world collaborative grasping settings.  
\end{abstract}

\IEEEpeerreviewmaketitle

\section{INTRODUCTION}

Task-driven agents often need to decide how to act based on partial and noisy state estimations which may greatly compromise performance.
We consider settings in which an agent is tasked with grasping an object based on a probabilistic estimation of its pose. Before attempting the grasp, another agent may assist by performing a sensing action and sending its observation to the grasping agent. 
In our settings of interest, sensing and communication may be costly or limited and there is a need to support the decision of which observation to perform by offering principled ways to assess the expected benefit.

\begin{figure}[t]
\centering
\includegraphics[width=0.49\textwidth]{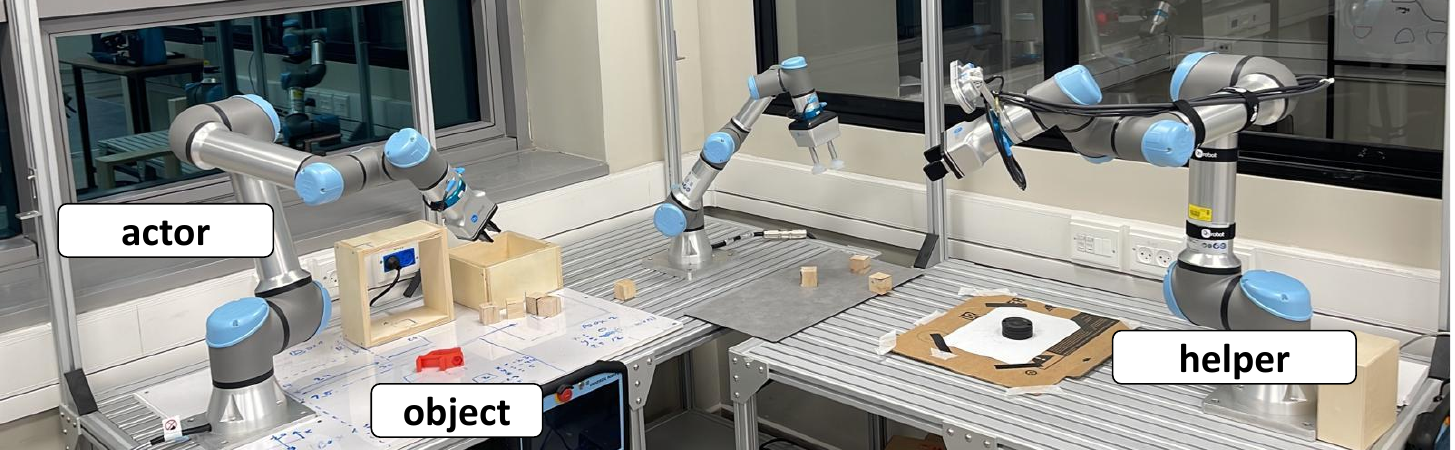}
\caption{Collaborative Grasping Example. }
\label{fig:example-setup}
\end{figure}

To demonstrate, consider the simplified automated manufacturing setting depicted in Figure \ref{fig:example-setup}. One agent, denoted as the {\em \actor}, is a robotic arm with a parallel-jaw gripper that is tasked with grasping an object (here, an adversarial object from  \cite{dexnet}). After a successful grasp, the object drops unexpectedly. Since the \actor~does not have a functioning sensor it can attempt to grasp the object based only on its estimation of the current position of the object. Alternatively, it can attempt the grasp after receiving an observation from another agent, the {\em \helper}, that is equipped with a functioning sensor (here, an OnRobot 2.5D Vision System).
The question that we pose is whether the \helper~can provide valuable assistance and what is the best position for the \helper~from which to provide the sensor reading among the possible options. Of course, the same question arises in single-agent settings in which it is the agent itself that needs to decide whether to perform a costly sensing action.

Beyond this illustrative example, grasping is an essential task for a wide range of robotic applications, including industrial automation, household robotics, agriculture, and more \cite{10.5555/3546258.3546288,DBLP:conf/icra/BicchiK00}. Accordingly, research on effective grasping capabilities has resulted in many solution approaches that can be generally divided into two main categories \cite{Sahbani2012AnOO,DBLP:conf/icra/BicchiK00}. In analytical approaches, a representation of the physical and dynamical models of the agent and the object are used when choosing a configuration from which to attemp a grasp \cite{hands1983kinematic,PrattichizzoTrinkle2008,PokornyKragicGrasp,berenson2011task}. In contrast, data-driven approaches rank labeled samples to come up with grasping policies. The ranking is usually based on a heuristic or on experiences collected from simulated or real robots \cite{6672028,Balasubramanian2012PhysicalHuman,PintoGupta2016Supersizing,NAHAVANDI2024102221,khansari2020action}.


We assume the \actor~is associated with a procedure for choosing a grasp given its \emph{belief} which represents its knowledge about the position of the object. To support the decision of which observation would be most beneficial, we formulate {\em value of assistance} (\VOA) {\em for grasping} and offer ways to compute it for estimating the benefit a sensing action will have on the probability of a successful grasp. This involves accounting for how the \actor's estimation will change based on the acquired observation and assessing how this change will affect the \actor's decision of how to attempt the grasp.

\VOA~is used to assess the informative value an observation will have and is therefore closely related to the well-established notions of {\em value of information} (VOI) and {\em information gain} (IG)~\cite{4082064,RUSSELL1991361,ZLtr9635,stachniss2005information,cai2009information}, which are widely used across multiple AI frameworks to assess the impact information will have on agents decisions and expected utility. We adapt these ideas to robotic settings. While in \cite{amuzig2023} we used \VOA~for assessing the effect localization information would have on a navigating robot's expected cost, here we use it in the context of a grasping task.

Perhaps closest to our work is {\em active perception} and {\em sensor planning}
\cite{9780576,pairet2018uncertainty,chandrasekhar2006localization,indelman2015planning, INDELMAN2013721,tian2022kimera,10.1007/10720246_21,Becker2009,ijcai2020p36} which refer to the integration of sensing and decision-making processes within a robotic system. This involves actively acquiring and utilizing information from the environment and selecting viewpoints or trajectories likely to reveal relevant information or reduce uncertainty \cite{TAYLOR2021102576,bajcsy2018revisiting}. While these include work on active perception in manipulation tasks they mostly focus on assessing the effect various perspectives will have on the ability to correctly locate and classify objects \cite{9780576,le2008active}. We offer a general formulation of \VOA~for grasping and novel ways to estimate the effect an observation will have on the probability of accomplishing a grasp.

Since our focus is on settings in which information acquisition actions are limited and may be performed by another agent, our work is also highly related to \emph{decision-theoretic communication} in particular, where agents communicate over a limited-bandwidth channel and messages are chosen to maximize the utility or effectiveness of the communication \cite{fern2014decision, Becker2009,10.1007/10720246_21,10.1145/860575.860598,Marcotte2020-wo}.
Our novelty is in offering measures that account for the manipulation and sensing capabilities of robotic agents when assessing the value of communicating an observation within a collaborative grasping setting. Our key contributions are the following: 
\begin{enumerate}
    \item We introduce and formulate {\em Value of Assistance} (\VOA) for grasping.
\item We instantiate \VOA~for a collaborative grasping setting with a robotic arm equipped with a gripper and another agent equipped with either a lidar or a depth camera. 
    \item We empirically demonstrate in both simulated and real-world robotic settings how \VOA~predicts the effect an observation will have on performance and how it can be used to identify the best assistive action.   
\end{enumerate}

\begin{figure}[t]
\centering
\medskip
\begin{subfigure}{0.1168\textwidth}
\centering
\includegraphics[trim = 25mm 5mm 25mm 5mm, clip, width=1\textwidth]{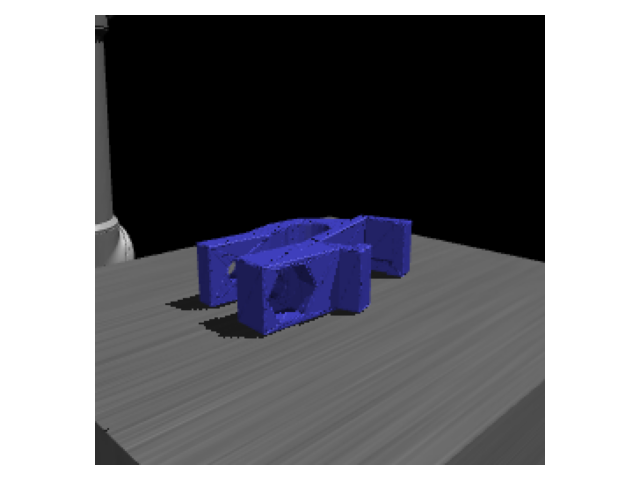}
\caption{}
\label{fig:example-obj-configs_a}
\end{subfigure}    
\begin{subfigure}{0.1168\textwidth}
\centering
\includegraphics[trim = 25mm 5mm 25mm 5mm, clip, width=1\textwidth]{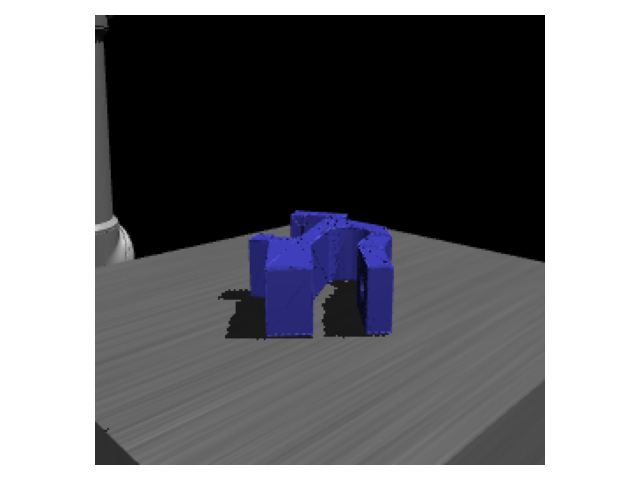}
\caption{}
\label{fig:example-obj-configs_b}
\end{subfigure}  
\begin{subfigure}{0.1168\textwidth}
\centering
\includegraphics[trim = 25mm 5mm 25mm 5mm, clip, width=1\textwidth]{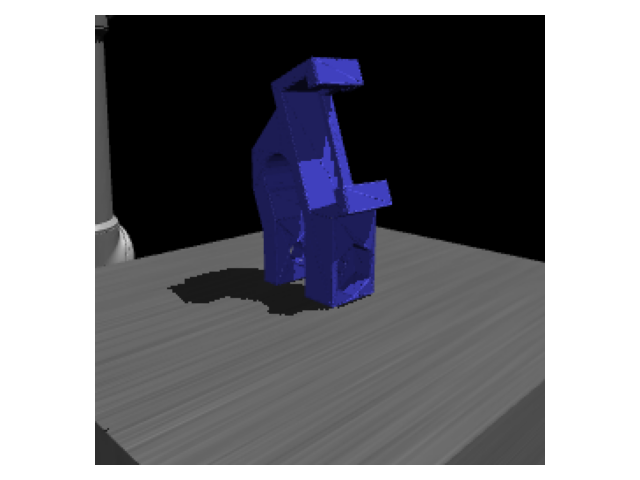}
\caption{}
\label{fig:example-obj-configs_c}
\end{subfigure}    
\begin{subfigure}{0.1168\textwidth}
\centering
\includegraphics[trim = 25mm 5mm 25mm 5mm, clip, width=1\textwidth]{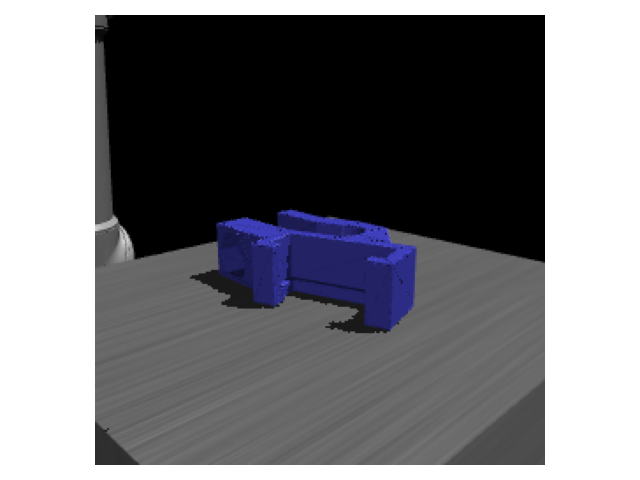}
\caption{}
\label{fig:example-obj-configs_d}
\end{subfigure}    
\caption{Example stable poses.}
\label{fig:example-obj-configs}
\end{figure}
\begin{figure}[t]
\centering
\begin{subfigure}{0.1168\textwidth}
\centering
\includegraphics[trim = 25mm 0mm 25mm 0mm, clip, width=1\textwidth]{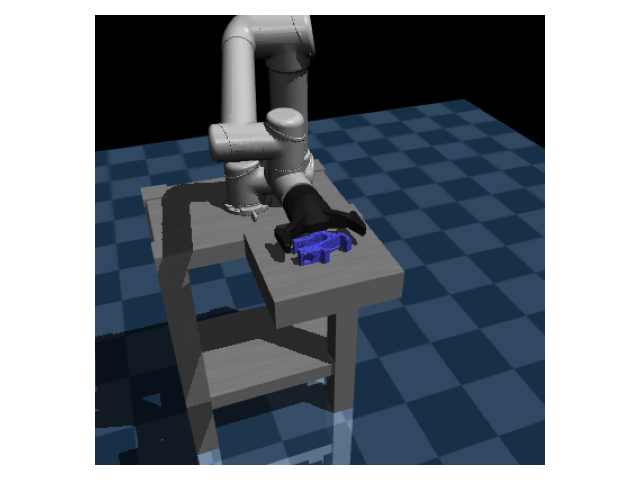}
\caption{}
\label{fig:example-grasp-configs_a}
\end{subfigure}
\begin{subfigure}{0.1168\textwidth}
\centering
\includegraphics[trim = 25mm 0mm 25mm 0mm, clip,width=1\textwidth]{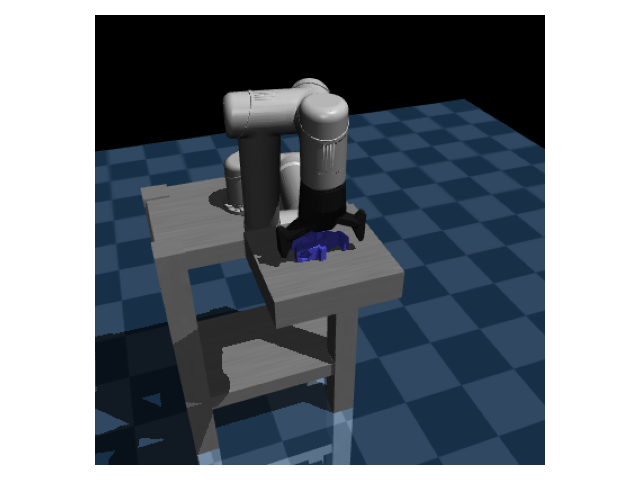}
\caption{}
\label{fig:example-grasp-configs_b}
\end{subfigure}
\begin{subfigure}{0.1168\textwidth}
\centering
\includegraphics[trim = 25mm 0mm 25mm 0mm, clip,width=1\textwidth]{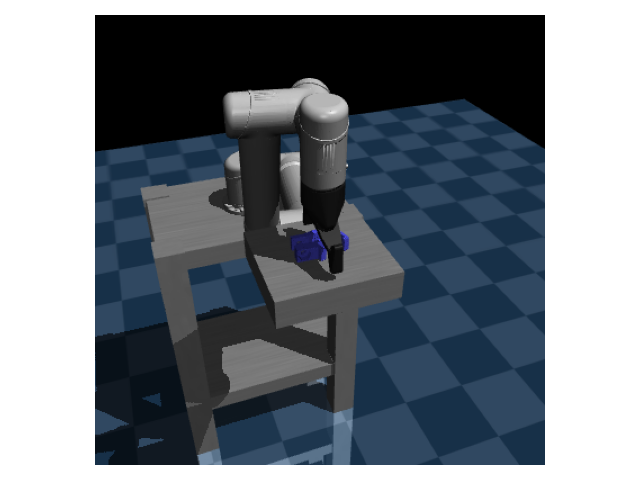}
\caption{}
\label{fig:example-grasp-configs_c}
\end{subfigure}
\begin{subfigure}{0.1168\textwidth}
\centering
\includegraphics[trim = 25mm 0mm 25mm 0mm, clip,width=1\textwidth]{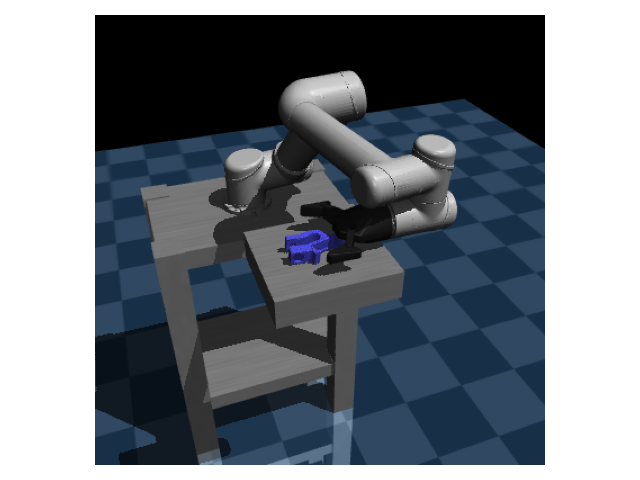}
\caption{}
\label{fig:example-grasp-configs_d}
\end{subfigure}
\caption{Example grasp configurations from which the \actor~can attempt to grasp the object - each configuration is associated with a score, i.e., probability of success.}
\label{fig:sim-example-grasp-configs}
\end{figure}

\begin{figure}[ht]
\centering
\medskip
\begin{subfigure}{0.1168\textwidth}
\centering
\includegraphics[trim = 2mm 4mm 6mm 4mm, clip, width=1\textwidth]{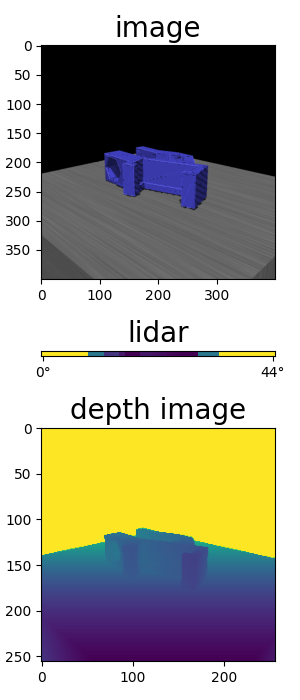}
\caption{}
\label{fig:example-obj-configs_a}
\end{subfigure}    
\begin{subfigure}{0.1168\textwidth}
\centering
\includegraphics[trim = 2mm 4mm 6mm 4mm, clip, width=1\textwidth]{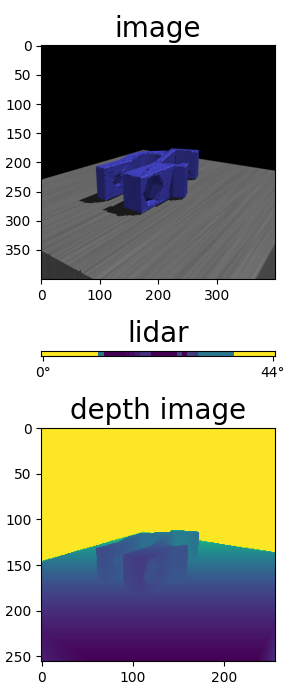}
\caption{}
\label{fig:example-obj-configs_b}
\end{subfigure}  
\begin{subfigure}{0.1168\textwidth}
\centering
\includegraphics[trim = 2mm 4mm 6mm 4mm, clip, width=1\textwidth]{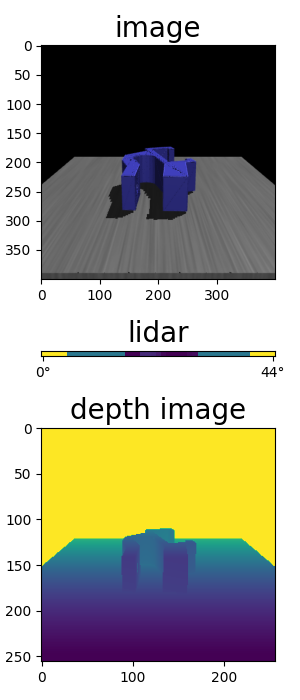}
\caption{}
\label{fig:example-obj-configs_c}
\end{subfigure}    
\begin{subfigure}{0.1168\textwidth}
\centering
\includegraphics[trim = 2mm 4mm 6mm 4mm, clip, width=1\textwidth]{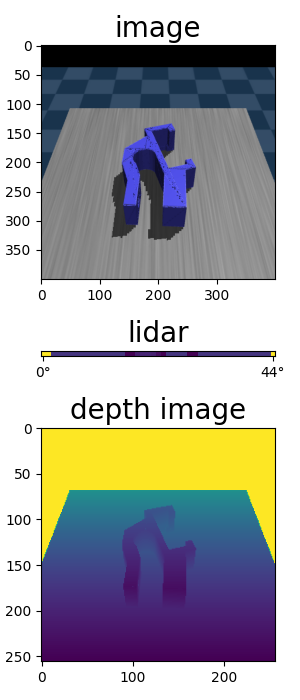}
\caption{}
\label{fig:example-obj-configs_d}
\end{subfigure}

\caption{Example sensor configurations. Each column represents the RGB image [top] lidar reading [middle]  and depth image [bottom] for a sensor configuration-object pose pair.}
\label{fig:example-sense-configs1}
\end{figure}

\section{Preliminaries} \label{sec: VOA}

To support a grasping task, where the object is assumed to be in a stable pose, we use a function that assigns a score to a grasping configuration - object stable pose pair.
\begin{definition}[Grasp Score]
Given a set of object poses $\allObjStablePoses$ and a set of grasp configurations $\graspConfigs$, 
a {\em grasp score function} 
$\graspScoreFunc: \graspConfigs \times \allObjStablePoses \mapsto [0,1]$
specifies the probability that an \actor~applying grasp configuration $\graspConfig\in \graspConfigs$ will successfully grasp an object at pose $\objPose \in \allObjStablePoses$,
i.e. 
$
\graspScoreFunc\left({\objPose, \graspConfig}\right) = P\left({ s \vert \objPose, \graspConfig}\right)
$, where $s$ is the event of a successful grasp. 
\end{definition}

The grasp score function may be evaluated analytically, by considering diverse factors such as contact area, closure force, object shape, and friction coefficient \cite{Murray1994AMI, DBLP:conf/icra/BicchiK00} or empirically, by using data-driven approaches such as \cite{mahler2017dex} where a deep learning model is trained to predict the quality of grasps based on depth images of the objects.


The \actor's choice of a grasp configuration relies on a {\em pose belief} which describes the perceived likelihood of each object pose within the set of possible stable poses $\objStablePoses$.

\begin{definition}[Pose Belief]\label{def:poseBel}
A {\em pose belief} $\poseBelief: \objStablePoses \mapsto [0,1]$ is a probability distribution over $\objStablePoses$.
\end{definition}

The pose belief is affected by different factors, including the model of the object and its dynamics and the collected sensory information. We formulate the initial belief after the object is dropped (Figure \ref{fig:example-setup}) using a joint probability model that captures the prior probability of stable poses, a von Mises PDF for the angle \cite{mardia2009directional}, and a multivariate normal distribution for the position on the plane. 
See Section \ref*{appendix:pose} of our online appendix for the formulation\footnote{\url{https://github.com/CLAIR-LAB-TECHNION/GraspVOA}}.

\begin{definition}[Expected Grasp Score] 
\label{def:MaxExpectedGraspScore}
For grasp configuration $\graspConfig$ and pose belief $\poseBelief$, the {\em \expectedGraspScore} $\graspScoreExpectation(\graspConfig, \poseBelief)
=
\mathbb{E}_{\objPose \sim \poseBelief}[\graspScoreFunc\left({ \graspConfig, \objPose }\right)]$
is the weighted aggregated grasp score over the set of possible poses. 
A \emph{maximal grasp} of $\poseBelief$, denoted $\graspConfigMax(\poseBelief)$, maximizes the expected grasp score, i.e., 
$\graspConfigMax(\poseBelief) = \argmax_{\graspConfig\in \graspConfigSet}{\graspScoreExpectation\left({\graspConfig, \poseBelief}\right)}$.
\end{definition}

The \actor~receives an {\em observation} from the \helper~which corresponds to its readings from a specific sensor configuration $\helperConfig \in \allHelperConfigs$ and object pose $\objPose \in \allObjStablePoses$. 
Our formulation of the {\em observation space} $\observations$ is general and represents the set of readings that can be made by the sensor that is available in the considered setting. In our evaluations, we used a planar \lidar~sensor for which a reading is an array of non-negative distances per angle $\mathbb{R}^{360}$ and a depth camera which emits a  
2D array $\mathbb{R}^{w \times h}$, where $w \times h$ are the image dimensions. 

As is common in the literature, (e.g., \cite{thrun2005probabilistic,lauri2022partially}), we consider obtaining an observation as a stochastic process.

\begin{definition}[Sensor Function]
\label{def:obsFunc} Given object pose $\objPose \in \allObjStablePoses$, sensor configuration $\helperConfig \in \allHelperConfigs$, and observation $\observation \in \observations$, sensor function $\sensorFunc\left({\observation,\objPose, \helperConfig}\right) = 
P\left({\observation \vert \objPose, \helperConfig }\right)$ 
provides the conditional probability of obtaining $\observation$ from $\helperConfig$ for $\objPose$.
\end{definition}

Notably, an agent may not be aware of the actual distribution and may instead only have a  
\emph{predicted sensor function}  
$\estimatedSensorFunc$ and a
\emph{predicted observation probability} $\observationPerceivedProbability\left({\observation \vert \objPose, \helperConfig}\right)$, based on a distribution which may be incorrect or inaccurate.






When receiving an observation $\observation$, the \actor~updates its belief using its \emph{belief update function} which defines the effect an observation has on the pose belief.

\begin{definition}[Belief Update]\label{def:beliefUpdate}
A belief update function \\
$\beliefUpdateFunc: \allPoseBeliefs \times \allObservations \times \allHelperConfigs \mapsto \allPoseBeliefs$ maps belief $\poseBelief\in \allPoseBeliefs$, observation $\observation \in \allObservations$ and sensor configuration $\helperConfig \in \allHelperConfigs$ to an updated belief $\updatedBelief$.
\end{definition}


The literature is rich with approaches for belief update (e.g., \cite{thrun2005probabilistic,stachniss2005information,indelman2015planning,kurniawati2022partially}). We use a Bayesian filter such that for any  observation $\observation \in \allObservations$ taken from sensor configuration $\helperConfig \in \allHelperConfigs$, the updated pose belief $\updatedBelief\left(\objPose\right)$ for pose $\objPose \in \objStablePoses$ is given as
\begin{equation}
\label{eq:belief update}
\updatedBelief\left({\objPose}\right)
=
\dfrac{
\observationPerceivedProbability\left({\observation \vert \objPose, \helperConfig}\right)
\poseBelief\left({\objPose}\right)
}{
\int_{\objPose' \in \objStablePoses}
\observationPerceivedProbability\left({\observation \vert \objPose', \helperConfig}\right) \poseBelief\left({\objPose'}\right)
d\objPose'
}
\end{equation}

\noindent where
$\poseBelief\left({\objPose}\right)$ is the estimated probability that $\objPose$ is the object pose prior to considering the new observation $\observation$.

When $\estimatedSensorFunc$ and $\observationPerceivedProbability$ describe stochastic processes they can be directly used for belief update using Equation \ref{eq:belief update}. Sometimes, it may be useful to consider deterministic sensor functions where the conditional distribution of an observation is replaced by a deterministic mapping  $\estimatedSensorFunc: \allObjStablePoses \times \allHelperConfigs \mapsto \allObservations$. 
For example, a deterministic sensor model would assign the value of a cell in a \lidar~reading based on the predicted distance between the \lidar~and the object surface at a specific angle. A stochastic sensor model would sample from a Gaussian distribution with this value as the mean and the specified error margins as the standard deviation.

In some cases, such as when using a deterministic sensor model, a similarity score $\obsSimScore:\observations\times\observations\mapsto [0,1]$ is used to compare the predicted and received observations and to compose a valid distribution function for $\observationPerceivedProbability$:

\begin{equation}
    \observationPerceivedProbability(\observation|\pose, \helperConfig) = \frac{\obsSimScore(\estimatedSensorFunc(\objPose,\helperConfig), o)}{\int_{\objPose' \in \objStablePoses} \obsSimScore(\estimatedSensorFunc(\objPose',\helperConfig), o) d\objPose'}
\label{eq:sim_mertic_update}
\end{equation}

The literature is rich with ways to measure $\obsSimScore$, which may vary between applications and sensor types. See  Appendix \ref*{sec:similarty} for a description of several approaches including using MSE for assessing the similarity between \lidar~sensor readings and an SSIM-based measure \cite{wang2004image} for depth images. 

\section{Value of Assistance (\VOA) for Grasping}

We offer ways to assess the effect sensing actions will have on the probability of successfully grasping an object.
We formulate this as a two-agent collaborative grasping setting with an {\em \actor} that is tasked with grasping an object for which the exact pose is not known and a {\em\helper} that can move to a specific configuration within its workspace to collect an observation from its sensor and share it with the \actor. 

We note that we use a two-agent model since it clearly distinguishes between the grasping and sensing capabilities.
Depending on the application, this model can be used to support an active perception setting in which a single agent needs to choose whether to perform a manipulation or sensing action if it is capable of both.

The \actor~chooses a grasp configuration based on its {\em pose belief}  (Definition \ref{def:poseBel}). While the object's exact pose is unknown, its shape is given and it is assumed to be in a stable pose (see Figure \ref{fig:example-obj-configs}). 
The \actor~needs to choose a feasible {\em \graspConfiguration} $\graspConfig \in \graspConfigs$ from which a grasp will be attempted (see Figure \ref{fig:sim-example-grasp-configs}) based on its pose belief and grasp score function. The \helper~can choose among a set of {\em sensor configurations} $\helperConfig \in\allHelperConfigs$ that 
offer different points of view of the object (Figure \ref{fig:example-sense-configs1}) and a potentially different effect on the
\actor's
pose belief.

We aim to assess {\em Value of Assistance} (\VOA) for grasping as the expected benefit an observation performed from a sensor configuration will have on the \actor's probability of a successful grasp. 
Our perspective is that of the  \helper~and its decision of which sensing action to perform. Accordingly, we seek a way to estimate beforehand the effect an expected observation will have on the \actor's belief and on its choice of configuration from which to attempt the grasp.

Importantly, the \helper's belief may be different than that of the \actor. 
We denote the \helper~and \actor~pose beliefs as $\poseBeliefHelper$ and $\poseBeliefActor$, respectively. We note that when considering a single agent with sensing and grasping capabilities the computation remains the same, with $\poseBeliefActor \equiv \poseBeliefHelper$.

A key element in \VOA~computation is the expected difference between the utility of the \actor~with and without the intervention. Here, utility is the grasp score of the configuration chosen by the \actor~based on its belief. 

\begin{sloppypar}

\begin{definition}[Value of Assistance (\VOA) for Grasping]
Given \actor~belief $\poseBeliefActor \in \allPoseBeliefs$, \helper~belief $\poseBeliefHelper \in \allPoseBeliefs$, helper perceived sensor function $\observationPerceivedProbability_{\helperSymbol}$, sensor configuration $\helperConfig \in \allHelperConfigs$ and \actor~belief update function $\beliefUpdateFunc_{\actorSymbol}$,
{
\small
\begin{align}
&\VOAFunc_{\assistanceAction}(\helperBelief,\actorBelief) \eqdef\\
& \quad \quad \mathbb{E}_{\objPose \sim \helperBelief}
\left[
\mathbb{E}_{\predictedObs \sim \observationPerceivedProbability_{\helperSymbol}\left({\observation \vert \objPose, \helperConfig }\right)}
\left[
\graspScoreFunc(\graspConfigMax(\predictedUpdatedBelief_{\actorSymbol}),\objPose)
\right]
- \graspScoreFunc(\graspConfigMax(\actorBelief),\objPose))\right] \notag
\end{align}
}
where $\predictedObs$ is the predicted observation and $\predictedUpdatedBelief_{\actorSymbol}$ is the predicted \actor's belief after receiving $\predictedObs$ from sensor configuration $\helperConfig$ and updating its belief, i.e., $\predictedUpdatedBelief_{\actorSymbol} = \beliefUpdateFunc_{\actorSymbol} (\poseBeliefActor, \predictedObs, \helperConfig)$.


\end{definition}
\end{sloppypar}
In the definition above, the maximal grasp $\graspConfigMax$ is the one that maximizes the expected grasp score $\graspScoreExpectation$ as in Definition \ref{def:MaxExpectedGraspScore}.
Since \VOA~estimation is performed by the \helper, observation $\predictedObs$ is extracted from its predicted observation probability $\observationPerceivedProbability_{\helperSymbol}$. In contrast, the \actor's belief update is based on $\beliefUpdateFunc_{\actorSymbol}$ and uses Equation \ref{eq:belief update} with its own perceived observation probability $\observationPerceivedProbability_{\actorSymbol}$.

For simplicity of presentation, we hereon assume that the \actor~and \helper~share an initial pose belief $\poseBelief$ before the grasp attempt. 
In addition, we assume the predicted observation $\predictedObs$ is generated using a deterministic sensor function such that
$\predictedObs=\estimatedSensorFunc(\objPose,\helperConfig)$. The updated belief is then a function of $\objPose$ and $\helperConfig$ and is denoted as $\predictedUpdatedBeliefDet$. We note that our evaluation and analysis can be adapted to the more general settings in which these assumptions are relaxed, but this allows us to use a simplified \VOA~formulation as follows  
\begin{equation}\label{eq:voasimple}
\VOAFunc_{\assistanceAction}(\poseBelief) 
\eqdef 
\mathbb{E}_{\objPose \sim \poseBelief}
\left[
\graspScoreFunc(\graspConfigMax(\predictedUpdatedBeliefDet),\objPose)
- \graspScoreFunc(\graspConfigMax(\poseBelief),\objPose)\right]
\end{equation}

Algorithm \ref*{alg:VOA2} in Appendix \ref*{sec:voaalg} describes how \VOA~can be used for supporting the \helper's decision of which sensing action to perform.   
The algorithm includes the ComputeVOA function for computing \VOA~for a sensor configuration. We also provide a complexity analysis for the algorithm.



\section{Empirical Evaluation}\label{sec: Empirical Evaluation}
The objective of our evaluation is to examine the ability of our proposed \VOA~measures to predict the effect sensing actions will have on the probability of a successful grasp and on finding one that maximizes this probability. 
With this objective in mind, our evaluation is comprised of three parts. 
\begin{enumerate}
    \item {\bf Evaluating \graspScore:} measuring the success ratio $
\graspScoreFunc\left({\objPose, \graspConfig}\right)$ for grasping an object at pose $\objPose$ from grasp configuration $\graspConfig$ for different objects.
    \item {\bf Evaluating
$\estimatedSensorFunc$:} examining the difference between the predicted sensor function $\estimatedSensorFunc$ and the readings of the actual sensor $\sensorFunc$. 
    \item {\bf Assessing \VOA}: assessing how well \VOA~estimates the effect observations will have on grasp score and its ability to identify the best sensing action. 
    

\end{enumerate}
\subsection{Experimental Setting}

\begin{figure}
    \centering
    \smallskip
    \includegraphics[width=0.45\textwidth]{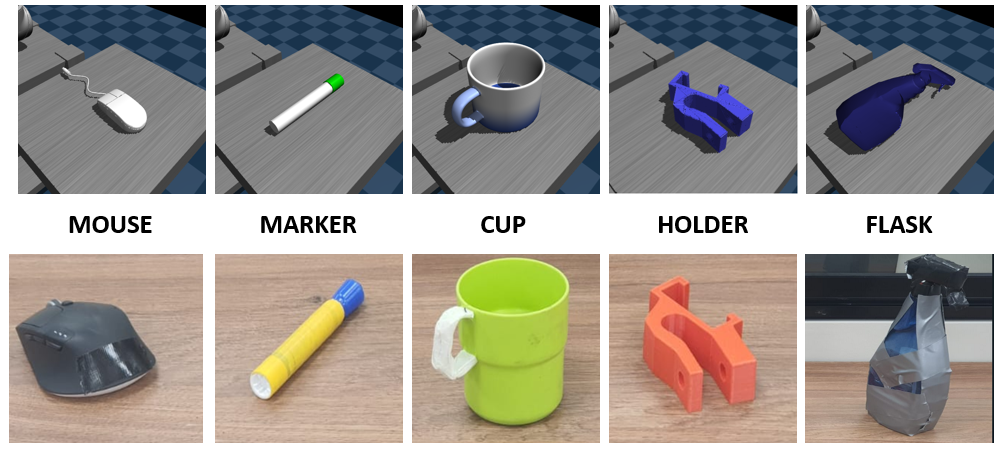}
    \caption{Evaluation objects}
    \label{fig:objects}
\end{figure}

We performed our evaluation in a two-agent robotic setting, both at the lab and in simulation. 
In our lab setting, depicted in Figure \ref{fig:example-setup},
the \actor~is a UR5e robotic arm~\cite{ur} 
with an OnRobot 2FG7 parallel jaw gripper \cite{onrobot}.
We used two implementations for the \helper~with two different sensors that might be available, depending on the setting: a LDS-01 \lidar~\cite{robotis} that could be moved on the x-y plane and a  2.5D Onrobot vision system \cite{onrobot} mounted on an adjacent UR5e arm. 
For simulation, we used a MuJoCo \cite{mujoco}
environment (depicted in Figure \ref{fig:sim-example-grasp-configs}) \cite{PaulDanielML}.
We simulated a \lidar~sensor using the MuJoCo depth camera, taking only one row of the camera's readings.
The simulated gripper was a Robotiq 2F-85 parallel jaw \cite{robotiqg}.
We used five objects for the simulation and lab experiments (Figure \ref{fig:objects}).
Objects meshes are based on the Dex-Net dataset \cite{dexnet}.

\begin{figure}[t]
\centering
\medskip
\begin{subfigure}{0.100\textwidth}
\centering
\includegraphics[trim = 25mm 5mm 17mm 5mm, clip, width=1\textwidth]{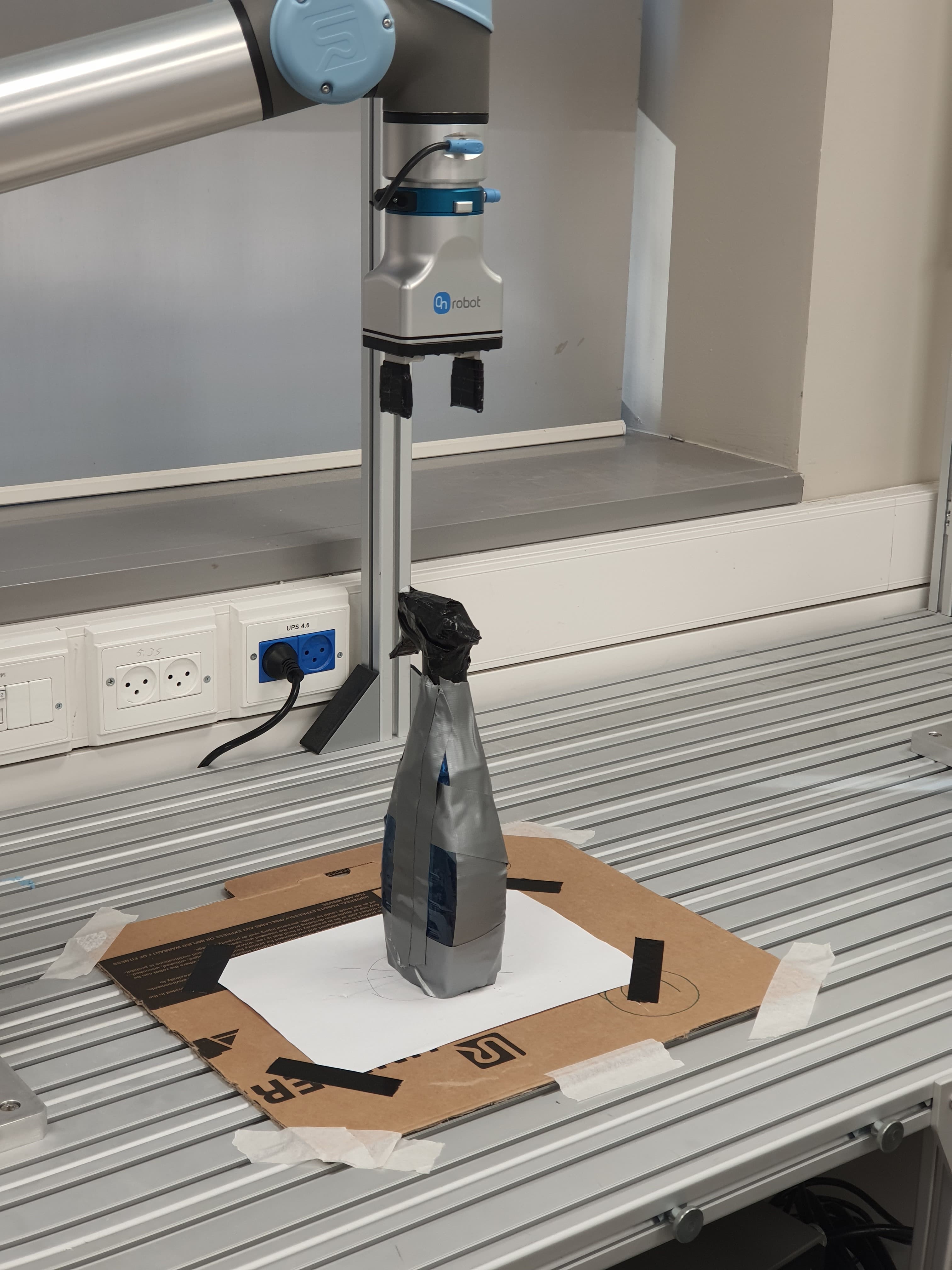}
\caption{}
\label{fig:example-obj-configs_a}
\end{subfigure}    
\begin{subfigure}{0.10\textwidth}
\centering
\includegraphics[trim = 25mm 5mm 17mm 5mm, clip, width=1\textwidth]{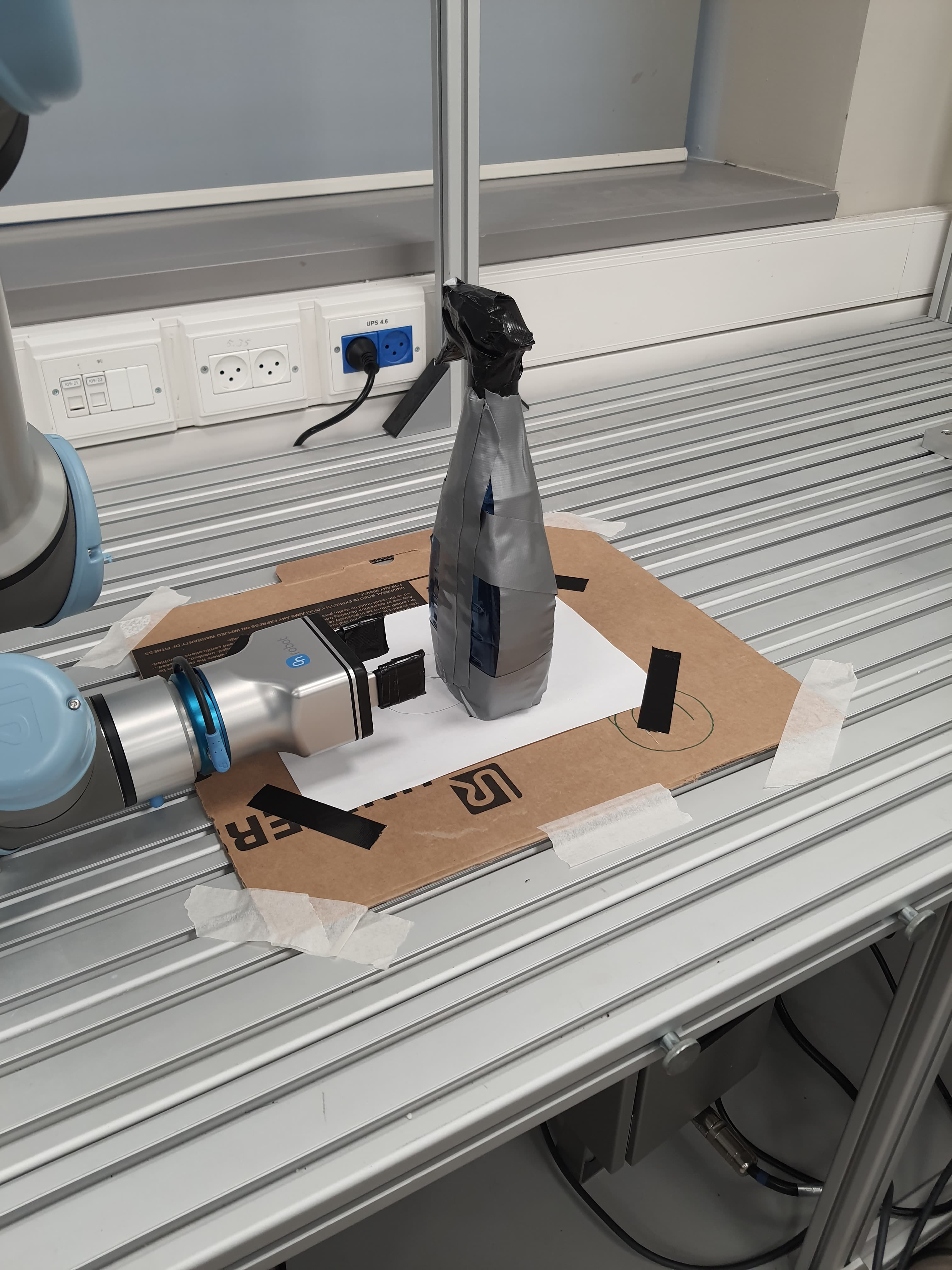}
\caption{}
\label{fig:example-obj-configs_b}
\end{subfigure}  
\begin{subfigure}{0.100\textwidth}
\centering
\includegraphics[trim = 25mm 5mm 17mm 5mm, clip, width=1\textwidth]{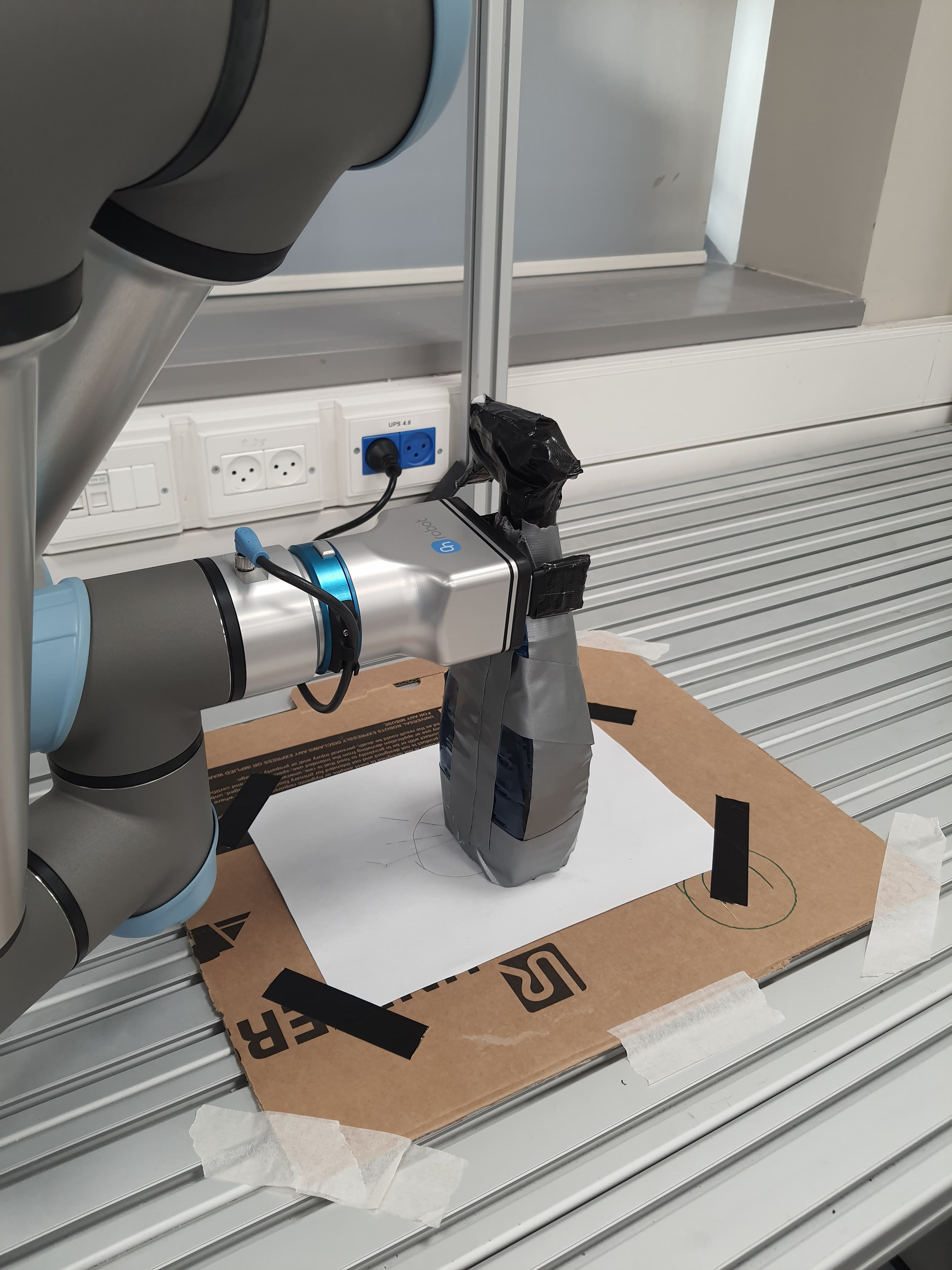}
\caption{}
\label{fig:example-obj-configs_c}
\end{subfigure}    
\begin{subfigure}{0.100\textwidth}
\centering
\includegraphics[trim = 25mm 5mm 17mm 5mm, clip, width=1\textwidth]{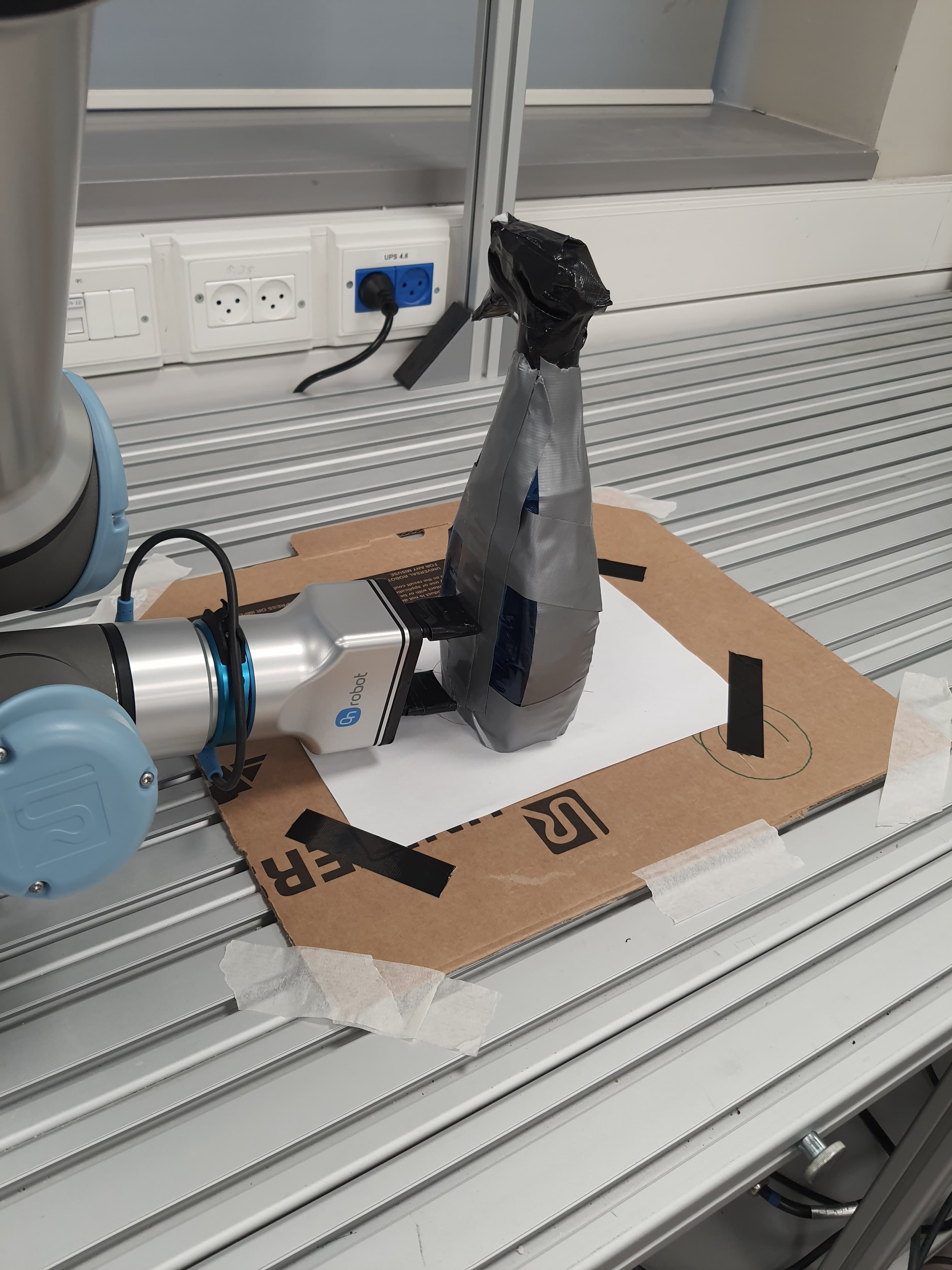}
\caption{}
\label{fig:flask_grasps}
\end{subfigure}

\caption{Four grasp configurations for FLASK at the lab}
\label{fig:lab-example-grasp-configs}
\vspace{-2mm} 
\end{figure}

We sampled a set $\objStablePosesSampled \subseteq \objStablePoses$ of stable poses and considered four possible grasps $\graspConfig$ indexed $1-4$ (see Figure \ref{fig:lab-example-grasp-configs}). 
We used both the \lidar~and depth camera. For each object pose $\objPose\in \objStablePosesSampled$, we recorded the observation $\observation$ and the predicted observation $\predictedObs$ received from each sensor configuration, indexed $I-IV$ for the \lidar~and $I-VI$ for the depth camera.

We empirically evaluated \graspScore~by examining a set of pose-grasp pairs for each object in both simulated and lab settings. Due to space constraints, the full details and results can be found in Section \ref*{sec:graps_score_eval} of our online appendix.

\begin{figure}
     \centering
      \begin{subfigure}{0.14\textwidth}
         \centering
         \includegraphics[trim = 30mm 15mm 50mm 25mm, clip, width=\textwidth]{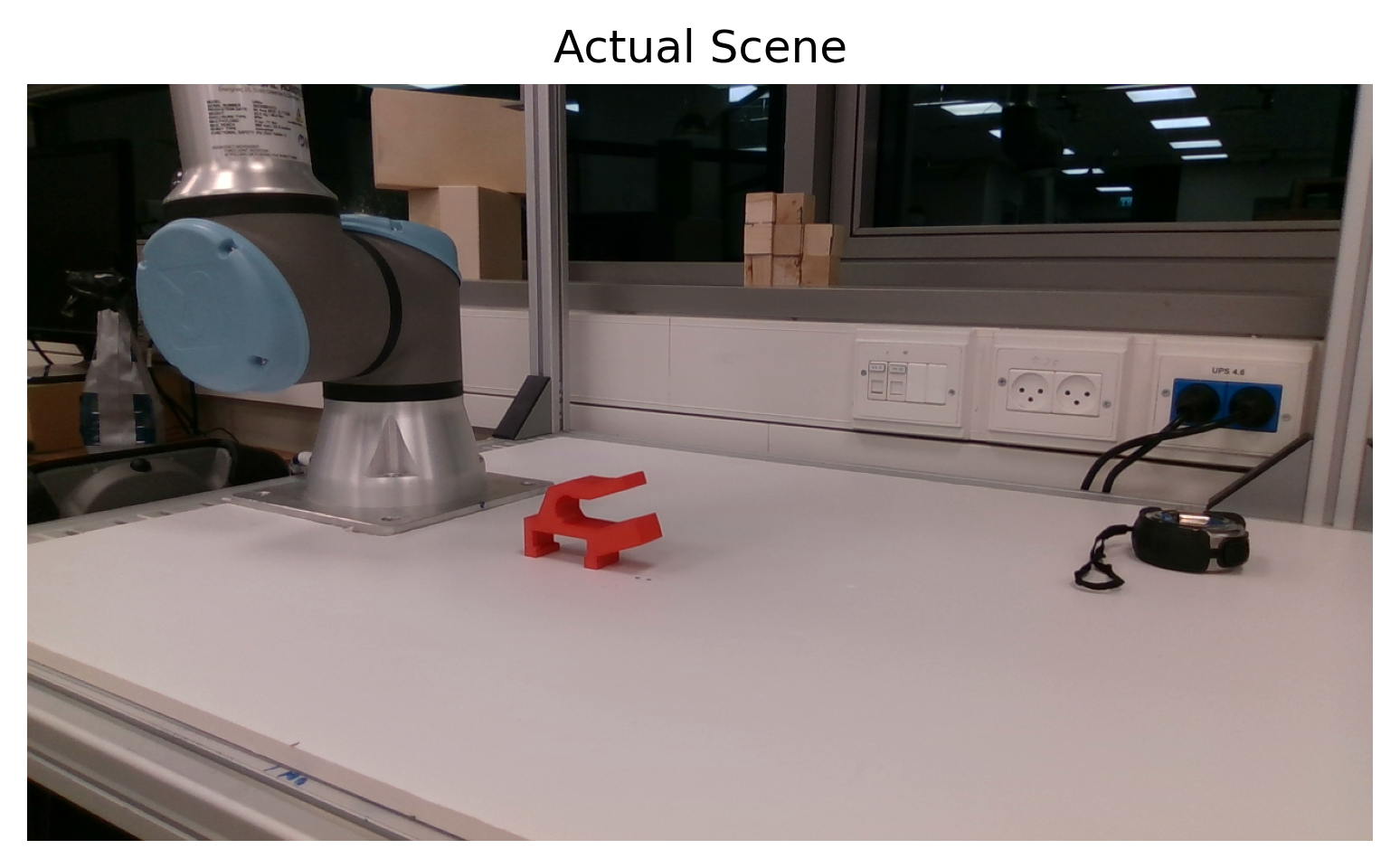}
         \label{fig:actual_scene}
         \caption{}
     \end{subfigure}~ 
     \begin{subfigure}{0.14\textwidth}
         \centering
         \includegraphics[trim = 30mm 15mm 50mm 25mm, clip,width=\textwidth]{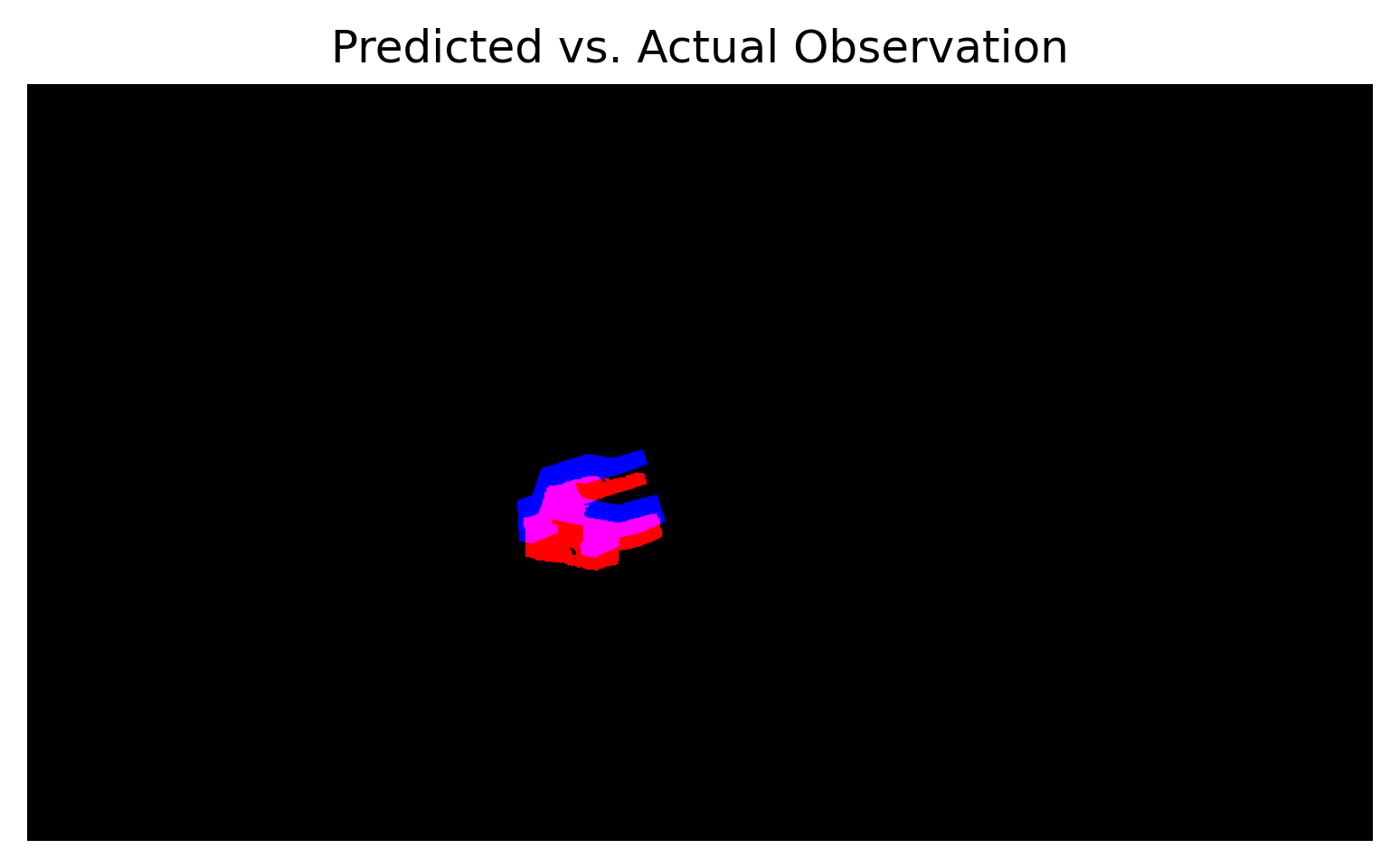}
         \label{fig:gen_vs_lab}
         \caption{}
    \end{subfigure}
    \begin{subfigure}{0.18\textwidth}
         \centering
         \includegraphics[trim = 0mm 3mm 0mm 14mm, clip,width=\textwidth]{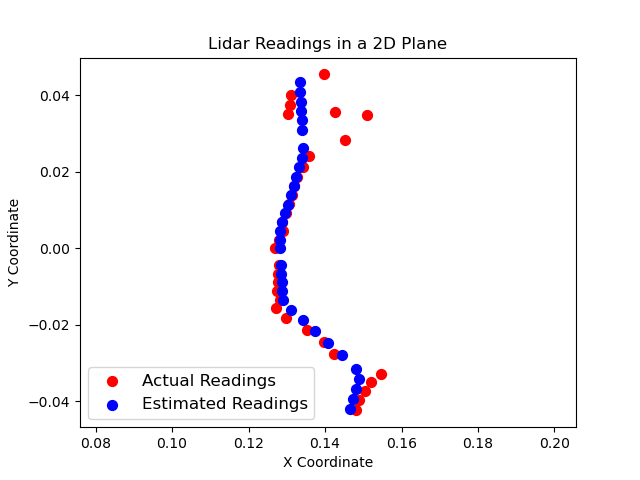}
         \label{fig:Lidar}
         \caption{}
     \end{subfigure}    
           
        \caption{Observations of HOLDER (a) Actual scene. (b) Comparing the predicted depth image (blue) and the lab-recorded image (red). (c) 
        Comparing the predicted (blue) and lab-recorded (red) 2D representation of the \lidar~reading.}
        \label{fig:o_depth-gen_lab}
            \vspace{-5mm}

\end{figure}

\subsection{Evaluating the Predicted Sensor Function
$\estimatedSensorFunc$}\label{sec:sensemp}
We evaluated our predicted sensor functions $\estimatedSensorFunc$ for both the \lidar~and depth camera by comparing their predicted observations to the readings collected from the sensor (Figure \ref{fig:o_depth-gen_lab}).
For each sensor configuration-object pose pair, we recorded the mean error of the difference between the measured and predicted readings.
For the \lidar, the set $\helperConfigs$ included four sensor configurations for each cardinal direction and $\estimatedSensorFunc$ computed the closest intersection between the simulated \lidar~ray and the object meshes for the relevant FoV. The actual observations generated by $\sensorFunc$ for this setup were collected in simulation and the lab. However, while in the lab the \lidar~was able to capture all objects, in simulation MOUSE and MARKER, could not be captured. 

For the depth camera, $\estimatedSensorFunc$ rendered synthetic images using the pyrender library \cite{pyrender2019} which involved projecting the 3D model of an object (transformed into a specific pose), onto a 2D plane using the camera's intrinsic and extrinsic parameters.
We evaluated our observation prediction accuracy compared to the actual image using the Intersection over Union (IoU) measure: we preprocessed the RBG images to extract the object masks and computed IoU between these and the corresponding synthetic images.
The set $\helperConfigs$  of sensor configurations was generated by randomly sampling 
robot configurations $q \in (-\pi, \pi]^6$ and translating them into camera poses using forward kinematics i.e. $\helperConfig = FK(q)$. Each sensor configuration $\helperConfig \in \helperConfigs$  was 
ranked using a heuristic 
\begin{equation}  
H(\helperConfig)=\left(1-\frac{D(\helperConfig)}{D_{max}}\right)+V(\helperConfig)\end{equation}
where $D(\helperConfig)$ is the Euclidean distance between the camera's center and the Point of Interest (PoI), representing the mean landing position after the object is dropped, $D_{max}$ is a maximum acceptable distance used for normalization, and $V(\helperConfig)$ is a visibility score, which assesses how centered the PoI is within the camera's FoV and is computed as the distance between the projection of the PoI onto the image plane and the center of the image divided by $R_{ref}$, a reference radius within the image plane that represents the boundary of acceptability.

\noindent{\bf Results:}
Table \ref{tab:obs_pred_eval} presents results per sensor for HOLDER (results for all objects are in our online appendix). For each sensor configuration $\helperConfig$ of the \lidar, the table shows the average, minimal and maximal error (Avg. Err., Min. Err. and Max. Err., respectively) in mm over object poses $\objStablePosesSampled$. Similarly, for the depth camera, we computed the average, maximal, and minimal IOU values. 

Results for the \lidar~show that the prediction errors are negligible given the dimensions of the objects examined. In contrast, for the depth camera, errors are more substantial with a maximal average of $0.6$. At the same time, 
results show varying performance across different configurations, depending on the object and its pose. For example, configuration $I$ gives high accuracy for some objects, while configuration $IV$ excels for others.

Inconsistencies between the observations are the result of several factors including the noise of the sensor itself, inconsistent scaling of the meshes with regard to the real objects, mismatches between objects and the meshes used for estimation, and inaccuracies in the placement of the objects in the lab (while the \observationprediction~is based on perfect object positioning). In addition, as the distance between the sensor and the object increases, the object occupies less of the sensor's FoV and the readings include fewer and less informative data points. Specific to the depth camera is the confusion caused by the reflection of bright light on shiny surfaces which may distort object shape.

Figure \ref{fig:o_depth-gen_lab} demonstrates inconsistencies between the predicted and actual observation of HOLDER at the lab. Here, this is due to the misplacement of the object. As we show next, despite these inconsistencies, the estimated observations are still useful for \VOA~computation.

\begin{table}[t]
    \centering
    \resizebox{0.45\textwidth}{!}{%
    \begin{tabular}{|c|c|c|c||c|c|c|c|}
        \hline  
        \multicolumn{4}{|c||}{\textbf{Lidar [mm]}} & \multicolumn{4}{c|}{\textbf{Depth Camera}} \\
        \hline
        $\helperConfig$ & Avg. & Min. & Max. & $\helperConfig$ & Avg. & Max & Min.  \\
         & Err. & Err. & Err & & IoU & IoU &  IoU  \\
        \hline
        $I$ & 1.6 & 0.7 & 3.2 & $I$ & 0.6 & 0.8 & 0.4 \\  
        $II$ & 3.9 & 0.7 & 4.8 & $II$ & 0.5 & 0.7 & 0.4 \\  
        $III$ & 0.8 & 0.5 & 1.0 & $III$ & 0.6 & 0.7 & 0.5 \\
        $IV$ & 3.0 & 0.6 & 4.7 & $IV$ & 0.6 & 0.7 & 0.6 \\ 
        - & - & - & - & $V$ & 0.4 & 0.6 & 0.3 \\ 
        - & - & - & - & $VI$ & 0.3 & 0.5 & 0.1\\ 
        \hline
    \end{tabular}%
    }
    \caption{Sensor prediction evaluation for the lab setting of HOLDER. For the \lidar~the lower values are better while for the depth camera higher values are better.}
    \label{tab:obs_pred_eval}
    \vspace{-5mm}
\end{table}


 


\subsection{Assessing \VOA}
We estimate the benefit of using \VOA~as a decision-making tool by assessing the benefit of choosing which sensor configuration to apply based on \VOA~values. Our evaluation uses the setting depicted in Figure \ref{fig:example-setup} and assumes the initial pose belief (after the object drops) is shared by the \actor~and \helper. The model of the initial belief is described in Section \ref{def:poseBel}.

We used three belief update functions per sensor, each based on a different similarity metric. 
For the lidar, $\beliefUpdateFunc_1$ uses a deterministic update rule that considers two observations $\observation_i, \observation_j$ as equivalent if for all angles the values are within a margin of $8$ mm, $\beliefUpdateFunc_2$ uses the similarity metric $
    \obsSimScore(\observation_i, \observation_j) = e^{-\left\Vert{ o_i - o_j}\right\Vert}
$ to update the belief based on Equation \ref{eq:sim_mertic_update}, while $\beliefUpdateFunc_3$ uses a multidimensional Gaussian over one observation while the other observation is the mean vector and the covariance matrix is the identity matrix.
For the depth camera, $\beliefUpdateFunc_4$ is based on the structure element of SSIM \cite{wang2004image}, $\beliefUpdateFunc_5$ is based on IoU between the two observations, and $\beliefUpdateFunc_6$ employs the cv2 library \cite{opencv_library} for contour matching to quantify the similarity between two masks by detecting their primary contours and comparing their shapes through a shape-matching algorithm.

\begin{figure}[h!]
    \centering
    \includegraphics[width=0.45\textwidth]{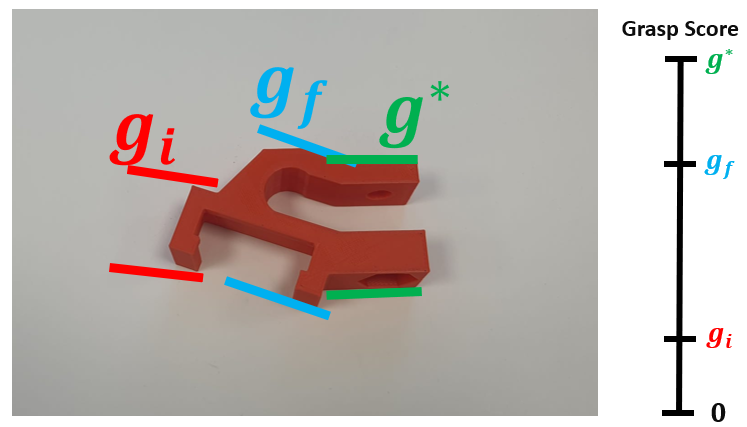}

    \caption{Grasps for HOLDER: best grasp $\graspConfig^*$, initial chosen grasp $\graspConfig_i$ and chosen grasp after the intervention $\graspConfig_f$.}
    \label{fig:three_grasps}
\end{figure}

\noindent{\bf Results:} Table \ref{tab:overall_voa} presents our evaluation for the different objects at the lab\footnote{due to space constraints, complete results as well as our implementation, are in our online appendix.}. For each setting, we consider three grasps: $\graspConfig^*$ is an optimal grasp, $\graspConfig_i$ is the grasp chosen by the \actor~based on its initial belief, and $\graspConfig_f$ is its post-intervention choice, i.e., after receiving an observation from a sensor configuration with the highest \VOA~value (see Figure \ref{fig:three_grasps}).  For each belief update function and object, the table reports the average values of:
\begin{itemize}
    \item  $\SCOREDIFF = \mathbb{E}_{\objPose \sim \poseBelief}\left[\graspScoreFunc(\graspConfig_f, \objPose)-\graspScoreFunc(\graspConfig_i, \objPose)\right]$: the weighted score difference between the chosen grasp after and before the intervention.
     \item $\BESTSCOREEXPRATIO=\frac{\mathbb{E}_{\objPose \sim \poseBelief}\left[\graspScoreFunc(\graspConfig_f, \objPose)-\graspScoreFunc(\graspConfig_i, \objPose)\right]}{\mathbb{E}_{\objPose \sim \poseBelief}\left[\graspScoreFunc(\graspConfig^*, \objPose)-\graspScoreFunc(\graspConfig_i, \objPose)\right]}$: the ratio between $\SCOREDIFF$~and the weighted score difference between the best configuration and the configuration chosen before the intervention.
     \item  $\ADVENTAGE = \frac{\BESTSCOREEXPRATIO(\helperConfig)}{E_{\helperConfig' \sim \helperConfigs}[\BESTSCOREEXPRATIO({\helperConfig'})]}$: the advantage of choosing the maximal \VOA~sensor configuration defined as the ratio between $\BESTSCOREEXPRATIO$ and the average over all configurations. This represents the difference between choosing a sensor configuration using \VOA~to choosing randomly.
\end{itemize}

\begin{figure}[h!]
    \centering
    \includegraphics[width=0.5\textwidth]{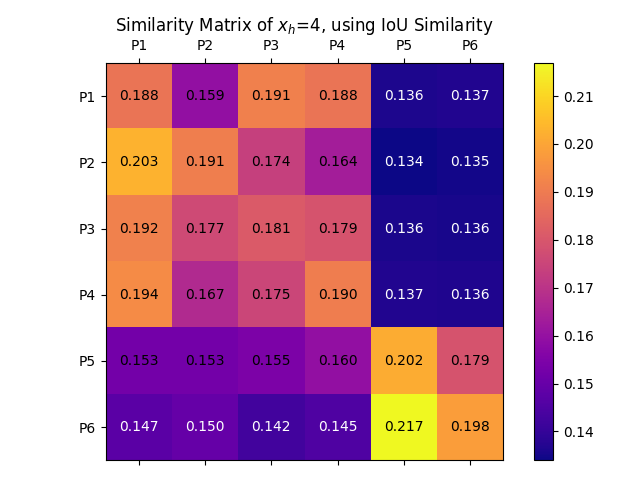}

    \caption{Similarity score between actual observation (rows) for each pose and predicted observations (columns). }
    \label{fig:iou_sim}
\vspace{-2mm} 
  
\end{figure}

\begin{figure}[h!]
\begin{subfigure}{0.075\textwidth}
\includegraphics[trim = 25mm 5mm 17mm 5mm, clip, width=1\textwidth]{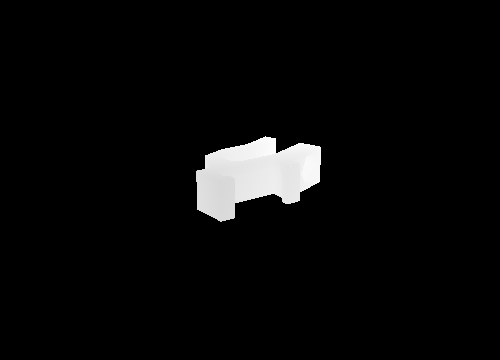}
\caption{P1}
\label{fig:example-obj-configs_a}
\end{subfigure}    
\begin{subfigure}{0.075\textwidth}
\includegraphics[trim = 25mm 5mm 17mm 5mm, clip, width=1\textwidth]{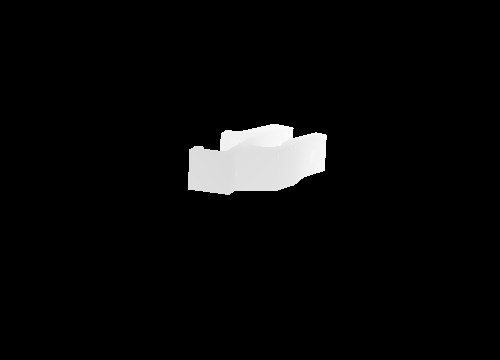}
\caption{P2}
\label{fig:example-obj-configs_a}
\end{subfigure}    
\begin{subfigure}{0.075\textwidth}
\includegraphics[trim = 25mm 5mm 17mm 5mm, clip, width=1\textwidth]{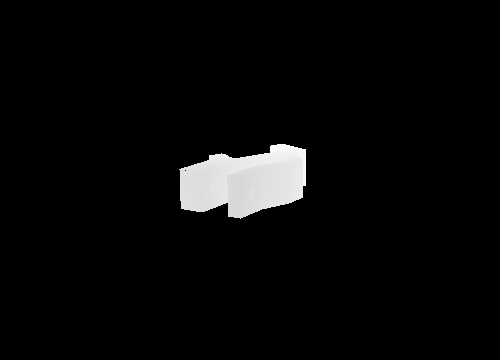}
\caption{P3}
\label{fig:example-obj-configs_b}
\end{subfigure}  
\begin{subfigure}{0.075\textwidth}
\includegraphics[trim = 25mm 5mm 17mm 5mm, clip, width=1\textwidth]{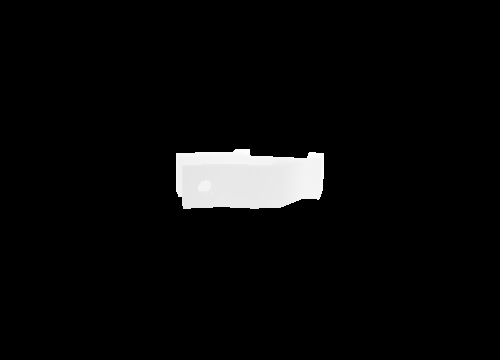}
\caption{P4}
\label{fig:example-obj-configs_c}
\end{subfigure}    
\begin{subfigure}{0.075\textwidth}
\includegraphics[trim = 25mm 5mm 17mm 5mm, clip, width=1\textwidth]{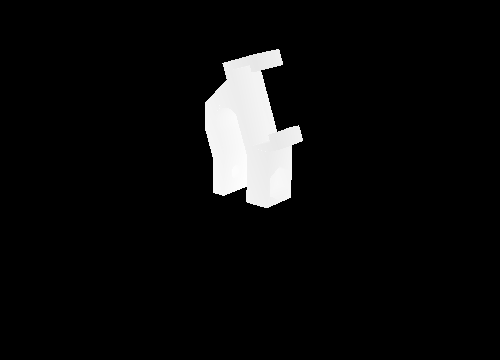}
\caption{P5}
\label{fig:holders_obs}
\end{subfigure}
\begin{subfigure}{0.075\textwidth}
\includegraphics[trim = 25mm 5mm 17mm 5mm, clip, width=1\textwidth]{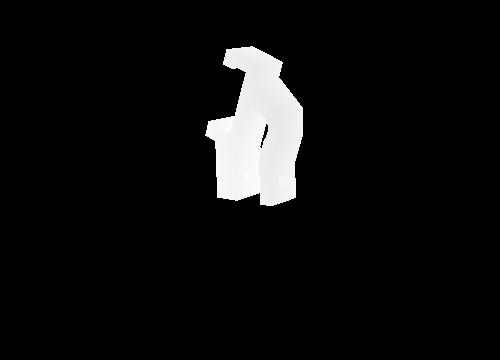}
\caption{P6}
\label{fig:example-obj-configs_a}
\end{subfigure}    

\caption{Observations for different object poses of HOLDER}
\label{fig:example-countour}
\vspace{-6mm} 

\end{figure}

\setlength{\arrayrulewidth}{0.2mm}
\setlength{\tabcolsep}{8pt}
\renewcommand{\arraystretch}{1.1}

\setlength{\arrayrulewidth}{0.2mm}
\setlength{\tabcolsep}{10pt}
\renewcommand{\arraystretch}{1.1}

\setlength{\arrayrulewidth}{0.2mm}
\setlength{\tabcolsep}{8pt}
\renewcommand{\arraystretch}{1.1}

\begin{table}[h!]
\centering
\begin{adjustbox}{angle=0}
\resizebox{0.47\textwidth}{!}{%
\begin{tabular}{|c|c|ccc|}
\specialrule{.1em}{.1em}{.2em} 
& &  $\SCOREDIFF$ & $\BESTSCOREEXPRATIO$ & $\ADVENTAGE$ \\
\specialrule{.2em}{.1em}{.2em} 
\multirow{6}{*}{$HOLDER$} 
& $\beliefUpdateFunc_1$ & {\bf 0.19} & 0.23 & 0.23 \\
& $\beliefUpdateFunc_2$  & 0.15 & 0.23 & {\bf 0.29} \\
& $\beliefUpdateFunc_3$  & 0.2 & {\bf 0.26} & 0.28\\
& $\beliefUpdateFunc_4$ & 0.03 & 0.15 & 0.13 \\
& $\beliefUpdateFunc_5$ & 0.0 & 0.0 & 0.0\\
& $\beliefUpdateFunc_6$ & 0.0 & 0.0 & 0.0 \\
\specialrule{.2em}{0em}{0em} 
\multirow{6}{*}{$EXPO$}
& $\beliefUpdateFunc_1$ & {\bf 0.29} & 0.31 & {\bf 0.29} \\
& $\beliefUpdateFunc_2$ & {\bf 0.29} & 0.31 & {\bf 0.29} \\
& $\beliefUpdateFunc_3$ & {\bf 0.29} & 0.31 & {\bf 0.29} \\
& $\beliefUpdateFunc_4$ & 0.04 & {\bf 0.37} & {\bf 0.29} \\
& $\beliefUpdateFunc_5$ & 0.00 & 0.00 & 0.00 \\
& $\beliefUpdateFunc_6$ & -0.00 & 0.29 & {\bf 0.29}\\
\specialrule{.2em}{0em}{0em} 
\multirow{6}{*}{$MOUSE$}
& $\beliefUpdateFunc_1$ & 0.00 & 0.06 & 0.04\\
& $\beliefUpdateFunc_2$ & 0.00 & 0.05 & 0.04\\
& $\beliefUpdateFunc_3$ & 0.00 & 0.07 & 0.00\\
& $\beliefUpdateFunc_4$ & {\bf 0.04}  & {\bf 0.44} & {\bf 0.53}\\
& $\beliefUpdateFunc_5$ & 0.00 & 0.00 & 0.00\\
& $\beliefUpdateFunc_6$ & 0.00 & 0.00 & 0.00\\
\specialrule{.2em}{0em}{0em} 
\multirow{6}{*}{$CUP$}
& $\beliefUpdateFunc_1$ & 0.30 & 0.33 & 0.30 \\
& $\beliefUpdateFunc_2$ & 0.30 & 0.33 & {\bf 0.41} \\
& $\beliefUpdateFunc_3$ & {\bf 0.37} & {\bf 0.63} & 0.37 \\
& $\beliefUpdateFunc_4$ & 0.24 & 0.38 & 0.27 \\
& $\beliefUpdateFunc_5$ & 0.00 & 0.00 & 0.00\\
& $\beliefUpdateFunc_6$ & 0.00 & 0.00 & 0.00\\
\specialrule{.2em}{0em}{0em} 
\multirow{6}{*}{$FLASK$}
& $\beliefUpdateFunc_1$ & {\bf 0.25} & {\bf 0.25} & {\bf 0.25} \\
& $\beliefUpdateFunc_2$ & {\bf 0.25} & {\bf 0.25} & {\bf 0.25} \\
& $\beliefUpdateFunc_3$ & {\bf 0.25} & {\bf 0.25} & {\bf 0.25} \\
& $\beliefUpdateFunc_4$ & 0.02 & {\bf 0.25} & {\bf 0.25} \\
& $\beliefUpdateFunc_5$ & 0.00 & 0.00 & 0.00 \\
& $\beliefUpdateFunc_6$ & 0.00 & 0.00 & 0.00 \\
\specialrule{.2em}{0em}{0em} 
\multirow{6}{*}{AVG}
& $\beliefUpdateFunc_1$ & 0.19 & 0.23 & 0.23 \\
& $\beliefUpdateFunc_2$ & 0.19 & 0.23 & 0.29 \\
& $\beliefUpdateFunc_3$ & {\bf 0.20} & {\bf 0.26} & 0.28 \\
& $\beliefUpdateFunc_4$ & 0.08 & 0.22 & {\bf 0.34} \\
& $\beliefUpdateFunc_5$ & 0.00 & 0.00 & 0.00 \\
& $\beliefUpdateFunc_6$ & 0.00 & 0.00 & 0.00 \\
\specialrule{.2em}{0em}{0em} 

\end{tabular}
}
\end{adjustbox}

\caption{Results per belief update function (best results per criteria are highlighted) }
\label{tab:overall_voa}
\vspace{-5mm} 

\end{table}

Results show that for all objects and for $4$ of our examined belief update functions ($\beliefUpdateFunc_1$-$\beliefUpdateFunc_4$), selecting the sensor configuration with the highest \VOA~value is beneficial in terms of the three examined measures.  
The smallest benefit is for MOUSE for which the initial grasp is optimal for all poses except one for which the optimal grasp has only a slight advantage. This subtlety is captured only by $\beliefUpdateFunc_4$.

Notably, belief update functions $\beliefUpdateFunc_5$ and $\beliefUpdateFunc_6$ which rely on IOU and contour matching, respectively, did not perform well on average for any of the objects. 
We associate this with the fact that the objects examined are small relative to their distance from the sensor, which is something these update functions are sensitive to.  Figure \ref{fig:iou_sim} presents the similarity scores between the actual (rows) and predicted (columns) observations for $\beliefUpdateFunc_5$ that relies solely on the IoU of the masks without considering depth values. This makes it hard for $\beliefUpdateFunc_5$ to differentiate between object poses that occupy the same area across multiple poses, as depicted in Figure \ref{fig:example-countour}.
The matrix shows a clear distinction between the standing positions $5$ and $6$ and the laying positions $1-4$, but the distinction within these two groups is challenging. 
Another critical issue is demonstrated by the values on the diagonal that show the low similarity scores between the predicted and actual observations. These indicate the sensitivity of $\beliefUpdateFunc_5$ to noise in the actual image, where even slight distortions can dramatically impact prediction quality. Similar results were observed for $\beliefUpdateFunc_6$ where noise dramatically distorts contours.

\section{CONCLUSION}

We introduced {\em Value of Assistance} (\VOA) for grasping and suggested ways to estimate it for sensing actions. Our experiments in both simulation and real-world settings demonstrate how our \VOA~measures predict the effect an observation will have on performance and how it can be used to support the decision of which observation to perform. 

Future work will account for optimization considerations of the \helper~and for integrating \VOA~in long-term and complex tasks. Another extension will consider multi-agent settings in which \VOA~can be used not only for choosing which assistive action to perform but also for choosing which agent to assist.





\bibliographystyle{IEEEtran}
\bibliography{bib}

\begin{thebibliography}{10}
\providecommand{\url}[1]{#1}
\csname url@samestyle\endcsname
\providecommand{\newblock}{\relax}
\providecommand{\bibinfo}[2]{#2}
\providecommand{\BIBentrySTDinterwordspacing}{\spaceskip=0pt\relax}
\providecommand{\BIBentryALTinterwordstretchfactor}{4}
\providecommand{\BIBentryALTinterwordspacing}{\spaceskip=\fontdimen2\font plus
\BIBentryALTinterwordstretchfactor\fontdimen3\font minus \fontdimen4\font\relax}
\providecommand{\BIBforeignlanguage}[2]{{%
\expandafter\ifx\csname l@#1\endcsname\relax
\typeout{** WARNING: IEEEtran.bst: No hyphenation pattern has been}%
\typeout{** loaded for the language `#1'. Using the pattern for}%
\typeout{** the default language instead.}%
\else
\language=\csname l@#1\endcsname
\fi
#2}}
\providecommand{\BIBdecl}{\relax}
\BIBdecl

\bibitem{dexnet}
``dexnet,'' \url{https://berkeleyautomation.github.io/dex-net}.

\bibitem{10.5555/3546258.3546288}
O.~Kroemer, S.~Niekum, and G.~Konidaris, ``A review of robot learning for manipulation: challenges, representations, and algorithms,'' \emph{J. Mach. Learn. Res.}, vol.~22, no.~1, jan 2021.

\bibitem{DBLP:conf/icra/BicchiK00}
A.~Bicchi and V.~Kumar, ``Robotic grasping and contact: {A} review,'' in \emph{Proceedings of the 2000 {IEEE} International Conference on Robotics and Automation, (ICRA)}.\hskip 1em plus 0.5em minus 0.4em\relax {IEEE}, 2000.

\bibitem{Sahbani2012AnOO}
A.~Sahbani, S.~El-Khoury, and P.~Bidaud, ``An overview of 3d object grasp synthesis algorithms,'' \emph{Robotics Auton. Syst.}, 2012.

\bibitem{hands1983kinematic}
A.~M. Hands, ``Kinematic and force analysis of,'' \emph{Journal of Mechanisms, Transmissions, and Automation in Design}, 1983.

\bibitem{PrattichizzoTrinkle2008}
D.~Prattichizzo and J.~C. Trinkle, ``Grasping,'' in \emph{Springer Handbook of Robotics}, B.~Siciliano and O.~Khatib, Eds.\hskip 1em plus 0.5em minus 0.4em\relax Springer, 2008.

\bibitem{PokornyKragicGrasp}
F.~T. Pokorny and D.~Kragic, ``Classical grasp quality evaluation: New algorithms and theory,'' in \emph{Proc. IEEE/RSJ Int. Conf. on Intelligent Robots and Systems (IROS)}.\hskip 1em plus 0.5em minus 0.4em\relax IEEE, Year, p. Page Range.

\bibitem{berenson2011task}
D.~Berenson, S.~Srinivasa, and J.~Kuffner, ``Task space regions: A framework for pose-constrained manipulation planning,'' \emph{The International Journal of Robotics Research}, 2011.

\bibitem{6672028}
J.~Bohg, A.~Morales, T.~Asfour, and D.~Kragic, ``Data-driven grasp synthesis—a survey,'' \emph{IEEE Transactions on Robotics}, vol.~30, no.~2, pp. 289--309, 2014.

\bibitem{Balasubramanian2012PhysicalHuman}
R.~Balasubramanian, L.~Xu, P.~D. Brook, J.~R. Smith, and Y.~Matsuoka, ``Physical human interactive guidance: Identifying grasping principles from human-planned grasps,'' \emph{IEEE Transactions on Robotics}, vol.~28, no.~4, pp. 899--910, 2012.

\bibitem{PintoGupta2016Supersizing}
L.~Pinto and A.~Gupta, ``Supersizing self-supervision: Learning to grasp from 50k tries and 700 robot hours,'' in \emph{Proc. IEEE Int. Conf. Robotics and Automation (ICRA)}, 2016.

\bibitem{NAHAVANDI2024102221}
\BIBentryALTinterwordspacing
S.~Nahavandi, R.~Alizadehsani, D.~Nahavandi, C.~P. Lim, K.~Kelly, and F.~Bello, ``Machine learning meets advanced robotic manipulation,'' \emph{Information Fusion}, vol. 105, p. 102221, 2024. [Online]. Available: \url{https://www.sciencedirect.com/science/article/pii/S1566253523005377}
\BIBentrySTDinterwordspacing

\bibitem{khansari2020action}
M.~Khansari, D.~Kappler, J.~Luo, J.~Bingham, and M.~Kalakrishnan, ``Action image representation: Learning scalable deep grasping policies with zero real world data,'' in \emph{2020 IEEE International Conference on Robotics and Automation (ICRA)}, 2020, pp. 3597--3603.

\bibitem{4082064}
R.~A. Howard, ``Information value theory,'' \emph{IEEE Transactions on Systems Science and Cybernetics}, vol.~2, no.~1, pp. 22--26, 1966.

\bibitem{RUSSELL1991361}
S.~Russell and E.~Wefald, ``Principles of metareasoning,'' \emph{Artificial Intelligence}, vol.~49, no.~1, pp. 361--395, 1991.

\bibitem{ZLtr9635}
S.~Zilberstein and V.~Lesser, ``Intelligent information gathering using decision models,'' Computer Science Department, University of Massachussetts Amherst, Tech. Rep. 96-35, 1996.

\bibitem{stachniss2005information}
C.~Stachniss, G.~Grisetti, and W.~Burgard, ``Information gain-based exploration using rao-blackwellized particle filters.'' in \emph{Robotics: Science and systems}, vol.~2, 2005, pp. 65--72.

\bibitem{cai2009information}
C.~Cai and S.~Ferrari, ``Information-driven sensor path planning by approximate cell decomposition,'' \emph{IEEE Transactions on Systems, Man, and Cybernetics, Part B (Cybernetics)}, vol.~39, no.~3, pp. 672--689, 2009.

\bibitem{amuzig2023}
A.~Amuzig, D.~Dovrat, and S.~Keren, ``Value of assistance for mobile agents,'' 2023.

\bibitem{9780576}
J.~Shin, S.~Chang, J.~Weaver, J.~C. Isaacs, B.~Fu, and S.~Ferrari, ``Informative multiview planning for underwater sensors,'' \emph{IEEE Journal of Oceanic Engineering}, vol.~47, no.~3, pp. 780--798, 2022.

\bibitem{pairet2018uncertainty}
{\`E}.~Pairet, J.~D. Hern{\'a}ndez, M.~Lahijanian, and M.~Carreras, ``Uncertainty-based online mapping and motion planning for marine robotics guidance,'' in \emph{2018 IEEE/RSJ International Conference on Intelligent Robots and Systems (IROS)}.\hskip 1em plus 0.5em minus 0.4em\relax IEEE, 2018, pp. 2367--2374.

\bibitem{chandrasekhar2006localization}
V.~Chandrasekhar, W.~K. Seah, Y.~S. Choo, and H.~V. Ee, ``Localization in underwater sensor networks: survey and challenges,'' in \emph{Proceedings of the ACM international workshop on Underwater networks}, 2006, pp. 33--40.

\bibitem{indelman2015planning}
V.~Indelman, L.~Carlone, and F.~Dellaert, ``Planning in the continuous domain: A generalized belief space approach for autonomous navigation in unknown environments,'' \emph{The International Journal of Robotics Research}, vol.~34, no.~7, pp. 849--882, 2015.

\bibitem{INDELMAN2013721}
\BIBentryALTinterwordspacing
V.~Indelman, S.~Williams, M.~Kaess, and F.~Dellaert, ``Information fusion in navigation systems via factor graph based incremental smoothing,'' \emph{Robotics and Autonomous Systems}, vol.~61, no.~8, pp. 721--738, 2013. [Online]. Available: \url{https://www.sciencedirect.com/science/article/pii/S092188901300081X}
\BIBentrySTDinterwordspacing

\bibitem{tian2022kimera}
Y.~Tian, Y.~Chang, F.~H. Arias, C.~Nieto-Granda, J.~P. How, and L.~Carlone, ``Kimera-multi: Robust, distributed, dense metric-semantic slam for multi-robot systems,'' \emph{IEEE Transactions on Robotics}, vol.~38, no.~4, 2022.

\bibitem{10.1007/10720246_21}
S.~Koenig and Y.~Liu, ``Sensor planning with non-linear utility functions,'' in \emph{Recent Advances in AI Planning}, S.~Biundo and M.~Fox, Eds.\hskip 1em plus 0.5em minus 0.4em\relax Berlin, Heidelberg: Springer Berlin Heidelberg, 2000, pp. 265--277.

\bibitem{Becker2009}
R.~Becker, A.~Carlin, V.~Lesser, and S.~Zilberstein, ``Analyzing myopic approaches for multi-agent communication,'' \emph{Computational Intelligence}, 2009.

\bibitem{ijcai2020p36}
R.~Mirsky, W.~Macke, A.~Wang, H.~Yedidsion, and P.~Stone, ``A penny for your thoughts: The value of communication in ad hoc teamwork,'' \emph{Proceedings of the Twenty-Ninth International Joint Conference on Artificial Intelligence}, pp. 254--260, July 2020.

\bibitem{TAYLOR2021102576}
\BIBentryALTinterwordspacing
A.~T. Taylor, T.~A. Berrueta, and T.~D. Murphey, ``Active learning in robotics: A review of control principles,'' \emph{Mechatronics}, vol.~77, p. 102576, 2021. [Online]. Available: \url{https://www.sciencedirect.com/science/article/pii/S0957415821000659}
\BIBentrySTDinterwordspacing

\bibitem{bajcsy2018revisiting}
R.~Bajcsy, Y.~Aloimonos, and J.~K. Tsotsos, ``Revisiting active perception,'' \emph{Autonomous Robots}, vol.~42, pp. 177--196, 2018.

\bibitem{le2008active}
Q.~V. Le, A.~Saxena, and A.~Y. Ng, ``Active perception: Interactive manipulation for improving object detection,'' \emph{Standford University Journal}, 2008.

\bibitem{fern2014decision}
A.~Fern, S.~Natarajan, K.~Judah, and P.~Tadepalli, ``A decision-theoretic model of assistance,'' \emph{Journal of Artificial Intelligence Research}, vol.~50, pp. 71--104, 2014.

\bibitem{10.1145/860575.860598}
C.~V. Goldman and S.~Zilberstein, ``Optimizing information exchange in cooperative multi-agent systems,'' in \emph{Proceedings of the Second International Joint Conference on Autonomous Agents and Multiagent Systems}, ser. AAMAS '03.\hskip 1em plus 0.5em minus 0.4em\relax New York, NY, USA: Association for Computing Machinery, 2003, p. 137–144.

\bibitem{Marcotte2020-wo}
R.~J. Marcotte, X.~Wang, D.~Mehta, and E.~Olson, ``Optimizing multi-robot communication under bandwidth constraints,'' \emph{Autonomous Robots}, vol.~44, no.~1, pp. 43--55, Jan. 2020.

\bibitem{Murray1994AMI}
R.~M. Murray, S.~Sastry, and Z.~Li, ``A mathematical introduction to robotic manipulation,'' 1994.

\bibitem{mahler2017dex}
J.~Mahler, J.~Liang, S.~Niyaz, M.~Laskey, R.~Doan, X.~Liu, J.~A. Ojea, and K.~Goldberg, ``Dex-net 2.0: Deep learning to plan robust grasps with synthetic point clouds and analytic grasp metrics,'' 2017.

\bibitem{mardia2009directional}
K.~V. Mardia and P.~E. Jupp, \emph{Directional statistics}.\hskip 1em plus 0.5em minus 0.4em\relax John Wiley \& Sons, 2009.

\bibitem{thrun2005probabilistic}
\BIBentryALTinterwordspacing
S.~Thrun, W.~Burgard, and D.~Fox, \emph{Probabilistic Robotics}, ser. Intelligent Robotics and Autonomous Agents series.\hskip 1em plus 0.5em minus 0.4em\relax MIT Press, 2005. [Online]. Available: \url{https://dl.acm.org/doi/10.5555/1121596}
\BIBentrySTDinterwordspacing

\bibitem{lauri2022partially}
M.~Lauri, D.~Hsu, and J.~Pajarinen, ``Partially observable markov decision processes in robotics: A survey,'' \emph{IEEE Transactions on Robotics}, vol.~39, no.~1, pp. 21--40, 2022.

\bibitem{kurniawati2022partially}
H.~Kurniawati, ``Partially observable markov decision processes and robotics,'' \emph{Annual Review of Control, Robotics, and Autonomous Systems}, vol.~5, pp. 253--277, 2022.

\bibitem{wang2004image}
Z.~Wang, A.~Bovik, H.~Sheikh, and E.~Simoncelli, ``Image quality assessment: From error visibility to structural similarity,'' \emph{Image Processing, IEEE Transactions on}, vol.~13, pp. 600 -- 612, 05 2004.

\bibitem{ur}
``{Universal Robotics-UR5},'' \url{https://www.universal-robots.com/products/ur5-robot/}.

\bibitem{onrobot}
``Onrobot,'' \url{https://onrobot.com/en/products/2fg7}.

\bibitem{robotis}
``Robotis,'' \url{www.robotis.us/360-laser-distance-sensor-lds-01-lidar}.

\bibitem{mujoco}
``mujoco,'' \url{https://mujoco.org/}.

\bibitem{PaulDanielML}
P.~Daniel, ``Mujoco\_rl\_ur5,'' \url{https://github.com/PaulDanielML/MuJoCo_RL_UR5}.

\bibitem{robotiqg}
``robotiq,'' \url{https://robotiq.com/products/2f85-140-adaptive-robot-gripper}.

\bibitem{pyrender2019}
\BIBentryALTinterwordspacing
mmatl, ``Pyrender,'' 2019, python 3D rendering toolkit. [Online]. Available: \url{https://github.com/mmatl/pyrender}
\BIBentrySTDinterwordspacing

\bibitem{opencv_library}
G.~Bradski, ``{The OpenCV Library},'' \emph{Dr. Dobb's Journal of Software Tools}, 2000.

\bibitem{Kolmogorov1933}
K.~Andrey, \emph{Foundations of the Theory of Probability}.\hskip 1em plus 0.5em minus 0.4em\relax New York: Chelsea Publishing Company, 1933.

\end{thebibliography}

\newpage
\clearpage
\setcounter{section}{0}

\section*{APPENDIX}
\section{Pose Belief}\label{appendix:pose}
The pose belief is affected by different factors, including the model of the object and its dynamics and collected sensory information. We formulate the initial belief after the object is dropped (Figure \ref{fig:example-setup}) using a joint probability model that captures the prior probability of stable poses, a von Mises PDF for the angle\cite{mardia2009directional}, and a multivariate normal distribution for the position on the plane.

Formally,

$\poseBelief(\objPose=(C, \theta, X, Y))=P(C) \cdot f_{\theta}(\theta | C; \mu_{\theta, C}, \kappa_C) \cdot f_{XY}(X, Y | C; \mu_C, \Sigma_C)$

In our model, the mean of the Gaussian distribution, $\mu_C$, represents the point where the agent drops the object, directly linking the expected landing position to the drop location. The covariance matrix, $\Sigma_C$, reflects the uncertainty in the object's landing position, which is influenced by the height from which the object is dropped; a higher drop height introduces greater variability in the landing outcome. The orientation parameter, $\mu_{\theta, C}$, is affected by the object's pose just before the drop, incorporating the initial orientation's influence on the final resting orientation.

\begin{itemize}
    \item $P(C)$ is the probability of the category $C$, where $C$ is a discrete random variable representing the category of stable poses for a given object. The variable $C$ takes values from a finite set of predefined categories, each corresponding to a distinct stable pose of the object. For example, the adversarial object in our example has five distinct categories, corresponding to the sides with a large enough surface.
    \item $f_{\theta}(\theta | C; \mu_{\theta, C}, \kappa_C) $ is a von Mises probability distribution function \cite{mardia2009directional} that represents the conditional probability density function of the angle $\theta$, given the category $C$, with $\mu_{\theta, C}$ as the mean direction and $\kappa_C$ as the concentration parameter specific to category $C$.
    \item $f_{XY}(X,Y;\mu,\Sigma)$ denotes the conditional probability density function of the position $(X,Y)$, given the category $C$, with $\mu_C$ as the mean position and $\Sigma_C$ as the covariance matrix, again specific to category $C$.
\end{itemize}

The \actor~uses its pose belief $\poseBelief$ and grasp score function $\graspScoreFunc$ to select a grasping configuration $\graspConfig\in \graspConfigSet$ from which to attempt to grasp the object.
A reasonable choice is to select a configuration that maximizes the \emph{expected grasp score}.

\section{Finding maximal \VOA}\label{sec:voaalg}
{
\small
\begin{algorithm}[h!]
\SetKwFunction{MEAN}{Mean}
\SetKwFunction{SAMPLE}{Sample}
\SetKwFunction{COMPUTEVOA}{ComputeVOA}
\SetKwProg{FN}{Function}{:}{}
\caption{Choose helping action with \VOA}
\label{alg:VOA2}
\KwIn{Grasp configuration set $\graspConfigSet$}
\KwIn{Sampled subset of help configurations $\helperConfigs \subseteq \allHelperConfigs$}
\KwIn{Actor belief $\poseBeliefActor$.}
\KwIn{Helper belief $\poseBeliefHelper$.}
\KwIn{Grasp score function $\graspScoreFunc$.}
\KwIn{Observation prediction function $\estimatedSensorFunc$}
\KwIn{Sampled subset of object pose set $\tilde{\objStablePoses}\subseteq \objStablePoses$.}
\KwIn{Belief update function $\beliefUpdateFunc$}
\KwOut{Sensor configuration $\helperConfig^{*} \in {\helperConfigs}$}

Compute $\graspConfigMax(\poseBeliefActor)$; 
\Comment{Chosen grasp without help. Definition \ref{def:MaxExpectedGraspScore}}

Compute $\graspScoreExpectation(\graspConfigMax(\poseBeliefActor), \poseBeliefHelper)$; 
\Comment{Mean score without help}

$\helperConfig^{*} \gets Null$\; \Comment{Chosen helping action}

${\VOAFunc}^{*} \gets 0$\; \Comment{best \VOA}

\For{$\helperConfig \in {\helperConfigs}$}
{
    $\VOAFunc_{\helperConfig} \gets \COMPUTEVOA(\helperConfig, \Tilde{\objStablePoses}, \graspScoreExpectation(\graspConfigMax(\poseBeliefActor), \poseBeliefHelper), \poseBeliefActor, \poseBeliefHelper, \beliefUpdateFunc)$\;
    \If{$\VOAFunc_{\helperConfig} > {\VOAFunc}^{*}$}
    {
        ${\VOAFunc}^{*} \gets \VOAFunc_{\helperConfig}$\;
        $\helperConfig^{*} \gets \helperConfig$\;
    }
}
\KwRet{$\helperConfig^{*}$}\;

\FN{\COMPUTEVOA$(\helperConfig, \Tilde{\objStablePoses}, \graspScoreExpectation(\graspConfigMax(\poseBeliefActor), \poseBeliefHelper), \poseBeliefActor, \poseBeliefHelper, \beliefUpdateFunc)$}
{
$\mathbf{\Gamma_{\Tilde{\objStablePoses}}} \gets \emptyset$;
\Comment{Score per sampled pose after help}

\For{$\objPose \in \Tilde{\objStablePoses}$} 
{
    $\estimatedObservation \gets \estimatedSensorFunc\left({\objPose, \helperConfig}\right)$; \Comment{predicted observation for pose}
    
    $\actorBelief^{\observation,\helperConfig} \gets \beliefUpdateFunc\left(\poseBeliefActor, \estimatedObservation, \helperConfig \right)$;
    \Comment{Predicted belief. Eq. \ref{eq:belief update}}
    
    Compute $\graspConfigMax(\actorBelief^{\observation,\helperConfig})$; 
    \Comment{Grasp with help. Def. \ref{def:MaxExpectedGraspScore}}
    
    Compute $\graspScoreFunc(\graspConfigMax(\actorBelief^{\observation,\helperConfig}),\poseBeliefHelper)$; 
    \Comment{Predicted score}
    
    $\mathbf{\Gamma_{\Tilde{\objStablePoses}}} \gets \mathbf{\Gamma_{\Tilde{\objStablePoses}}} \cup \{ 
    \graspScoreFunc(\graspConfigMax(\actorBelief^{\observation,\helperConfig}),\poseBeliefHelper)
    \}$\;
}

\Return $\MEAN(\mathbf{\Gamma}_{\Tilde{\objStablePoses}}) - \graspScoreExpectation(\graspConfigMax(\poseBeliefActor), \poseBeliefHelper)$\;
}
\textbf{end function}
\end{algorithm}
}

We observe that $\estimatedSensorFunc\left({\objPose, \helperConfig}\right)$ could be a bottleneck in the algorithm, depending on the sensor used. We will denote its complexity as $O(\estimatedSensorFunc)$. Also it depends on the implementation, but $\omega$ can be a bottleneck.

Analysing the time complexity of \COMPUTEVOA:
\begin{enumerate}
    \item Outer Loop Over $\Tilde{\objStablePoses}$: The outer loop iterates over the sampled set of poses $\Tilde{\objStablePoses}$, so it runs $|\Tilde{\objStablePoses}|$ times.
    \item item Observation Prediction Function and Belief Update: Within each iteration of the outer loop, the $\estimatedSensorFunc$ function is called, adding $O(\estimatedSensorFunc)$. The belief update is also called, which contains its own internal loop over $\Tilde{\objStablePoses}$ and calls  $\estimatedSensorFunc$ and $\omega$ in each iteration. Therefore, the belief update contributes $O(|\Tilde{\objStablePoses}|(\estimatedSensorFunc+\omega))$.
    \item Grasp Calculation Over $\graspConfigSet$ and $\Tilde{\objStablePoses}$: The best grasp calculation, which occurs for each pose, has a complexity of $|\graspConfigSet||\Tilde{\objStablePoses}|$ since it runs for each grasp configuration and each pose.
\end{enumerate}

Combining the complexities from these components, we get:

    \begin{itemize}
        \item For each pose in $\Tilde{\objStablePoses}$, the estimated sensor function and the belief update yield $O(\estimatedSensorFunc) + |\Tilde{\objStablePoses}| \times O(\estimatedSensorFunc)$.
        \item The grasp calculation contributes $|\graspConfigSet||\Tilde{\objStablePoses}|$.
    \end{itemize}

So the total complexity is given by: $$|\Tilde{\objStablePoses}| \times \left(O(\estimatedSensorFunc) + |\Tilde{\objStablePoses}| \times O(\estimatedSensorFunc)+ |\Tilde{\objStablePoses}| \times O(\omega) + |\graspConfigSet||\Tilde{\objStablePoses}|\right)$$.

Simplifying this expression, we get: $$O\left(|\Tilde{\objStablePoses}|^2\left(|\graspConfigSet|+ \estimatedSensorFunc + \omega \right)\right)$$

In \ref{alg:VOA2} we iterate over $\helperConfigs$, where in each iteration we run \COMPUTEVOA, Thus, the total complexity of the algorithm is: $O\left(|\helperConfigs||\Tilde{\objStablePoses}|^2\left(|\graspConfigSet|+ \estimatedSensorFunc + \omega \right)\right)$

One significant optimization we've introduced involves the computation of $\estimatedSensorFunc$, which are utilized at least $|P_o|$ times within the algorithm. Recognizing this, we precalculate a similarity matrix denoted as $S$, which consists of all pairwise similarities between the estimated observations. In other words, $S_{i,j}=\omega(\observation_i,\observation_j)$. It's important to note that the observations are exclusively used for the purpose of calculating $\omega(\observation_i,\observation_j)$ in the belief update. The resulting complexity of \COMPUTEVOA is: $$O\left(|\Tilde{\objStablePoses}|^2\left(|\graspConfigSet|+ \omega\right) + |\Tilde{\objStablePoses}|\estimatedSensorFunc\right)$$

\section{Observation Similarity Scores}\label{sec:similarty}

The following discussion presents three ways to implement $\obsSimScore$ but we note that our framework is not restricted to these methods. 
The first measure deterministically decides whether two observations $\observation$ and $\observation'$ are equivalent, i.e.,  $ \observation \equiv \observation'$. We note that such a measure does not constrain the sensor to be deterministic but instead imposes a deterministic rule for deciding whether two observations are considered equivalent, e.g., if the difference between their values is within some bound. 

A second option is to use any arbitrary similarity metric as long as it simulates a probability in the sense that the Kolmogorov axioms \cite{Kolmogorov1933} are met when comparing one observation with the simulated probability distribution's mean.
For example, to assess structural similarity between depth image pairs $\observation$ and $\observation'$, we applied the structure element of Structural Similarity Index (SSIM) \cite{wang2004image} using the formula: $$s(\observation, \observation') = \frac{\sigma_{\observation\observation'} + c}{\sigma_{\observation}\sigma_{\observation'}+c}
$$

In this formula:
\begin{itemize}
    \item $\sigma_{x}$ and $\sigma_{y}$ are the standard deviation of the observations $\observation$ and $\observation'$.
    \item $\sigma_{\observation\observation'}$ is the covariance between observations $\observation$ and $\observation'$.
    \item $c$ is a small constant to stabilize the division.
\end{itemize}

As a third option, uncertainty is approximated using a compact parametric model, e.g., a Gaussian distribution, which is defined by two parameters: the mean $\mu$ and the variance $\sigma^2$. In the context of a similarity model, the mean could represent the point of maximum similarity (which could be zero distance for identical vectors), and the variance could control how quickly the similarity decreases as the distance increases. Here we assume that the similarity between two observations $\observation$ and $\observation'$ can be modeled as a function of the distance between them in some feature space. Using the Euclidean distance yields: $$\omega(\observation,\observation')=\exp\left(-\frac{\|\observation-\observation'\|}{2\sigma^2}\right)$$ Small $\sigma^2$ makes the similarity function sharply peak around zero distance, making the model very sensitive to small changes in distance and large $\sigma^2$ makes the decay of similarity with distance more gradual.

Examples of these methods of approximating the observation probability are given in our empirical evaluation in Section \ref{sec: Empirical Evaluation}.


Note that the observation prediction function  
is equivalent to a deterministic sensing function, as defined in Definition \ref{def:obsFunc}.
In the general case, the observation function can be stochastic, thus the grasp score after help is stochastic, and there is an expectation in the definition. In the case of a deterministic observation function, the expectation is not necessary.

Ideally, the helper agent would be able to compute the value difference for to support it's decision about a helping action, but in practice, the pose is unknown, thus the actual value difference is intractable. Instead, the helper agent can estimate the value difference using it's belief $\poseBeliefHelper$.

\section{Evaluating Grasp Score}\label{sec:graps_score_eval}
\subsection{Evaluating \graspScore} \label{sec:evalGraspScore}
\noindent{\bf Setup:}
We empirically evaluated \graspScore~by examining a set of pose-grasp pairs for each object. In the lab setting, we examined 6 poses and 4 grasps. In simulation, we examined 3-6 poses and 3-4 grasps. 
For every pose-grasp pair, we ran 100 and 15 grasp attempts in simulation and at the lab, respectively.
In simulation, for each grasp attempt a Gaussian noise was added to the object position (in meters) and orientation (in radians). We ran 4 experiment sets with different standard deviations of the noise: 0, 0.01, 0.03 and 0.05. At the lab,  
we used a set of 15 noisy variations of the object position and orientation. 

\noindent {\bf Results:}
For each pose-grasp pair, we counted the number of successful grasps. Results  demonstrate
the diversity of our setup, with some grasps that have a high probability of success for several poses and others that are adequate for a limited set of poses. 

\begin{figure}[h!]
     \centering
    \includegraphics[width=0.5\textwidth]{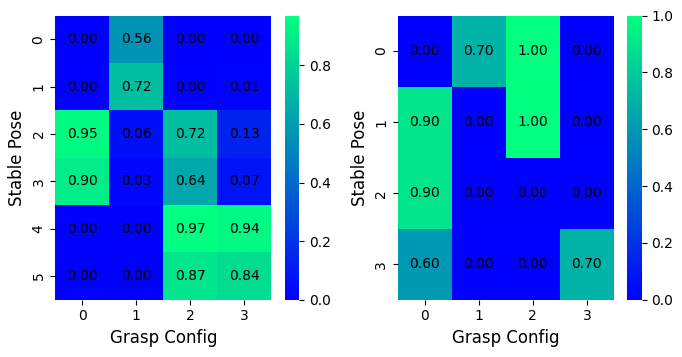}
    \caption{FLASK grasp score for  the simulated [left] and lab [right] settings.}
        \label{fig:flask graspscore}
    
\end{figure}

Figure \ref{fig:flask graspscore} shows the ratio of successful grasps for FLASK in the simulated and lab settings. Rows represent stable poses and the columns represent grasp configurations (results for all other objects are in the supplementary material). In simulation, with a uniform belief over the poses, the agent should attempt grasp 2 because it has the highest expected grasp score. If, however, the belief associates a high probability to pose 2 or 3, the agent should attempt grasp 0.

\end{document}


\maketitle
\thispagestyle{empty}
\pagestyle{empty}

\begin{function}
\caption{"ExpectedGraspScore($\poseBelief$, $\theta$, $\varphi$, $\objStablePoses$)"}
\label{alg:expectedScore}
\KwOut{$q$}
$q \gets 0$\;
\For{$\objPose \in \objStablePoses$}
{
$q 
\gets
q
+
\poseBelief\left({\objPose}\right)
\theta\left({ \varphi, \objPose }\right)
$\;
}
\end{function}

\begin{function}
\SetKwFunction{EGS}{ExpectedGraspScore}
\caption{"GraspArgMax($\poseBelief$, $\graspScoreFunc$, $\objStablePoses$, $\graspConfigSet$)"}
\label{alg:argmax}
\KwOut{$\graspConfigMax$}
$\graspConfigMax \gets \varnothing$\;
$q_{\max} \gets 0$\;
\For{$\graspConfig \in \graspConfigSet$}
{
$q \gets \EGS\left({\poseBelief, \graspScoreFunc, \graspConfig, \objStablePoses}\right)$\;
\If{$q_{\max} < q$}
{
$q_{\max} \gets q$\;
$\graspConfigMax \gets \graspConfig$\;
}
}
\end{function}

\begin{procedure}
\SetKwFunction{EGS}{ExpectedGraspScore}
\caption{"UpdateBelief($\poseBelief$, $\observationPerceivedProbability$, $\observation$, $\sensorConfig$, $\objStablePoses$)"}
\label{alg:beliefUpdate}
\KwOut{$\predictedPoseBelief$}
$q \gets 0$\;
\For{$\objPose \in \objStablePoses$}
{
$q 
\gets
q
+
\poseBelief\left({\objPose}\right)
\observationPerceivedProbability\left({\observation  \vert \objPose, \sensorConfig}\right)
$\;
}
\For{$\objPose \in \objStablePoses$}
{
$\predictedPoseBelief\left({\objPose}\right) \gets 
\frac{
\observationPerceivedProbability\left({\observation \vert \objPose, \sensorConfig}\right)
\poseBelief\left({\objPose}\right)
}{q}
$\;
}
\end{procedure}

\begin{algorithm}
\SetKwFunction{EGS}{ExpectedGraspScore}
\SetKwFunction{GAM}{GraspArgMax}
\SetKwFunction{UB}{UpdateBelief}
\caption{Calculate \VOA}
\label{alg:main}
\KwIn{Sensor Configuration $\sensorConfig$.}
\KwIn{Sensor Belief $\poseBeliefSensor$.}
\KwIn{Gripper Belief $\poseBeliefGripper$.}
\KwIn{Perceived Observation Probability $\observationPerceivedProbability$.}
\KwIn{Grasp Score Function $\graspScoreFunc$.}
\KwIn{Observation Generator Function $\sensorFunc$.}
\KwIn{Object Pose Set $\objStablePoses$.}
\KwIn{Grasp Configuration Set $\graspConfigSet$.}
\KwOut{u}
$\graspConfigMax \gets \GAM\left({\poseBeliefGripper, \graspScoreFunc, \objStablePoses, \graspConfigSet}\right)$\;
$q_{\graspScoreFunc} \gets 
\EGS\left({\poseBeliefGripper, \graspScoreFunc, \graspConfigMax, \objStablePoses}\right)$\;
$u \gets 0$\;
\For{$\objPose \in \objStablePoses$}
{
$\observation = \sensorFunc\left({\objPose, \sensorConfig}\right)$\;
$\predictedPoseBeliefGripper \gets \UB\left({\poseBeliefGripper, \observationPerceivedProbability, \observation, \sensorConfig, \objStablePoses}\right)$\;
$\graspConfigMax \gets \GAM\left({\predictedPoseBeliefGripper, \graspScoreFunc, \objStablePoses, \graspConfigSet}\right)$\;
$q \gets \EGS\left({\predictedPoseBeliefGripper, \graspScoreFunc, \graspConfigMax, \objStablePoses}\right)$\;
$
u \gets
u + q \cdot \poseBeliefSensor\left({\objPose}\right)
$\;
}
$u \gets u - q_{\graspScoreFunc}$\;
\end{algorithm}

\begin{table}[htbp]\label{tbl:obsEval}
\caption{Evaluating \observationPrediction$\text{ }$results}
\centering
\begin{tabular}{|c|c|c|}
\hline
\textbf{angle [degree]} & \textbf{mean [$m$]} & \textbf{variance [$m^2$]} \\
\hline
-8 & 0.0085 & 0.0008 \\
\hline
-7 & 0.0101 & 0.0007 \\
\hline
-6 & 0.0031 & 0.0000 \\
\hline
-5 & 0.0017 & 0.0000 \\
\hline
-4 & 0.0016 & 0.0000 \\
\hline
-3 & 0.0001 & 0.0000 \\
\hline
-2 & 0.0002 & 0.0000 \\
\hline
-1 & 0.0001 & 0.0000 \\
\hline
0 & -0.0001 & 0.0000 \\
\hline
1 & 0.0004 & 0.0000 \\
\hline
2 & 0.0009 & 0.0000 \\
\hline
3 & 0.0015 & 0.0000 \\
\hline
4 & 0.0018 & 0.0000 \\
\hline
5 & 0.0018 & 0.0000 \\
\hline
6 & 0.0001 & 0.0000 \\
\hline
7 & 0.0028 & 0.0006 \\
\hline
8 & 0.0005 & 0.0000 \\
\hline
\end{tabular}
\label{tbl:obsEval}
\end{table}

\addtolength{\textheight}{-12cm}   



\section*{APPENDIX}


\bibliographystyle{IEEEtran}
    
\bibliography{bib}